%% file: main.tex
\documentclass[letterpaper, 10 pt, conference]{ieeeconf}  
\usepackage{amsmath}
\usepackage{amssymb}
\usepackage{graphicx}
\usepackage{subcaption}
\usepackage{xcolor}
\usepackage{booktabs} 
\usepackage{tabularx}
\usepackage{multicol}
\usepackage{multirow}
\usepackage{hyperref}
\usepackage{algorithm}
\usepackage{algpseudocode}
\usepackage{circledsteps}
\usepackage{url}
\usepackage[normalem]{ulem}
\let\labelindent\relax
\usepackage{enumitem}
\usepackage{mathabx}

\algrenewcommand\algorithmicrequire{\textbf{Input:}}
\algrenewcommand\algorithmicensure{\textbf{Output:}}

\IEEEoverridecommandlockouts                              

\newcommand{\add}[1]{\textcolor{black}{#1}}

\title{\LARGE \bf
On-the-fly hand-eye calibration for the da Vinci surgical robot
}

\author{Zejian Cui$^{1}$ and Ferdinando Rodriguez y Baena$^{1}$
\thanks{*This work is supported by funding from the Department of Mechanical Engineering, Imperial College London}
\thanks{$^{1}$Both authors are with the Mechatronics in Medicine Laboratory, the Hamlyn Centre for Robotics Surgery, Department of Mechanical Engineering, Imperial College London, Exhibition Road, London, SW7 2AZ, UK
        {\tt\small \{zejian.cui19, f.rodriguez\}@imperial.ac.uk}}%
}

\begin{document}
\maketitle
\thispagestyle{empty}
\pagestyle{empty}

\newcommand{\Blue}[1]{\textcolor{blue}{#1}}
\newcommand{\Red}[1]{\textcolor{red}{#1}}
\newcommand{\Green}[1]{\textcolor{green}{#1}}

\newcommand{\vecfont}[1]{\texttt{\textbf{#1}}}
\newcommand{\matfont}[1]{\text{\textbf{#1}}}
\newcommand{\pd}[2]{\frac{\partial {#1}}{\partial {#2}}}
\newcommand{\ThetaVec}{\mathbf{\Theta}}
\newcommand{\tVec}{\vecfont{t}}

\newcommand{\Hcand}{\mathcal{\check{K}}}
\newcommand{\Hlist}{\mathcal{\check{E}}}
\newcommand{\Hfinal}{\mathcal{\hat{K}}_{final}}
\newcommand{\InnList}{\mathcal{P}_{inn}}
\newcommand{\ResList}{\mathcal{P}_{res}}

\newcommand{\xpre}{\hat{\vecfont{x}}^{(t-1)}}
\newcommand{\sigxpre}{\mathbf{\hat{\Sigma}_x}^{(t-1)}}
\newcommand{\xnow}{\hat{\vecfont{x}}^{(t)}}
\newcommand{\sigxnow}{\mathbf{\hat{\Sigma}_x}^{(t)}}
\newcommand{\xnowprior}{\check{\vecfont{x}}^{(t)}}
\newcommand{\sigxnowprior}{\mathbf{\check{\Sigma}_x}^{(t)}}
\newcommand{\sigenow}{\mathbf{\hat{\Sigma}_e}^{(t)}}
\newcommand{\sigvnow}{\mathbf{\hat{\Sigma}_v}^{(t)}}
\newcommand{\sigepre}{\mathbf{\hat{\Sigma}_e}^{(t-1)}}
\newcommand{\sigvpre}{\mathbf{\hat{\Sigma}_v}^{(t-1)}}

\newcommand{\etal}{\textit{et al.}}

\input{Files/Abstract}
\input{Files/Introduction}

\input{Files/Methods}

\input{Files/Experiments}
\input{Files/Results}
\input{Files/Discussions}
\input{Files/Conclusions}

\input{Files/Appendix}

\addtolength{\textheight}{-0cm}
\bibliographystyle{ieeetr}
\bibliography{Files/references} 
\end{document}

%% file: Files/Abstract.tex
\begin{abstract}
In Robot-Assisted Minimally Invasive Surgery (RMIS), accurate tool localization is crucial to ensure patient safety and successful task execution. However, this remains challenging for cable-driven robots, such as the da Vinci robot, because erroneous encoder readings lead to pose estimation errors. In this study, we propose a calibration framework to produce accurate tool localization results through computing the hand-eye transformation matrix on-the-fly. The framework consists of two interrelated algorithms: the feature association block and the hand-eye calibration block, which provide robust correspondences for key points detected on monocular images without pre-training, and offer the versatility to accommodate various surgical scenarios by adopting an array of filter approaches, respectively. To validate its efficacy, we test the framework extensively on publicly available video datasets that feature multiple surgical instruments conducting tasks in both \textit{in vitro} and \textit{ex vivo} scenarios, under varying illumination conditions and with different levels of key point measurement accuracy. The results show a significant reduction in tool localization errors under the proposed calibration framework, with accuracies comparable to other state-of-the-art methods while being more time-efficient. 

\end{abstract}

%% file: Files/Introduction.tex
\section{INTRODUCTION}

Hand-eye calibration is a well-established research topic in the field of robotics: it aims to find the homogeneous transformation matrix between the camera and robot base frame. When these two frames are closely aligned, the robot can accurately command its end-effector to a given pose indicated in the camera frame. Specifically, for medical robots, it is a common practice to incorporate a vision system, whether monocular or stereoscopic, allowing surgeons to react promptly during surgery according to real-time camera feedback. For a tele-operated surgical platform, such as the da Vinci robot (Intuitive Surgical Inc, Sunnyvale, California), surgeons operate on the surgeon-side console, observing patient-side instruments that mimic their hand movements, from the view of a stereo endoscopic camera. This calibration is usually conducted using the built-in set-up joints controllers (SUJs), both pre-operatively and intra-operatively, to ensure instrument movements are visually aligned with surgeons' hand movements, contributing towards an intuitive and ergonomic operation. However, it was reported \cite{d2021autonomous} that, the calibration result provided by SUJs may introduce errors that impair instrument's tip positioning accuracy. Although humans can automatically compensate for these errors during tele-operation, in cases where surgical subtasks automation, or human-robot interactions, are required, it is imperative to resolve these errors so that the instrument can reach the target position \cite{FRDAC_ZC}.

In addition to initial hand-eye calibration errors, tip positioning accuracy further suffers from inaccurate joint angle readings from the encoder, induced by the cable-driven mechanism \cite{dvrk_caveats}, which is intrinsic to the da Vinci Classic, and da Vinci Si systems. These encoder reading errors are time-varying, because of the wear and tear in cables and increasing non-linearities in pulley-cable friction. Richter \etal  \cite{UCSD_Unified} proposed a unified approach of lumping encoder reading errors into a complementary homogeneous transformation matrix, such that accurate tip positioning can be achieved through consistent estimations of this matrix. Alternatively, Hwang \etal \cite{Berkeley_automation_Hwang} turned to direct calibrations on encoder readings using a pre-trained neural network model. Their training-based approach relies on collecting ground truth joint angles using a bespoke marker, which can be time-consuming \textit{per se}. Additionally, a model trained on one instrument may not be applicable to another instrument, even of the same type, due to variations in transmission kinematics degeneracy sustained by each instrument. Considering the large variety of surgical instruments and their disposable nature, in this study, we posit that on-the-fly hand-eye calibration approach offers better generalization for improving instrument positioning accuracy. 



\subsection{Contributions}
In this work, we aim to provide a generalizable on-the-fly hand-eye calibration framework for a wide range of surgical platforms that is also practicable for adoption in a real clinical monocular setup. To this end, we have made the following contributions

\begin{enumerate}[label=\arabic*)]
    \item A training-free key point association algorithm that leverages analytical Jacobian matrices and is easily generalizable for different surgical instruments for which CAD models are available.
    \item A visibility check algorithm that accelerates the data association process by promptly removing invisible key points from the candidate list.
    \item A pool of hand-eye calibration methods that adopt filter-based approaches to allow for different noise distributions that arise in various surgical scenarios. 
\end{enumerate}

It should be noted that, to make the framework more generalizable, key point detection methods are outside the scope of this study. It is assumed that either traditional or learning-based methods can be adopted. Although they may result in different levels of measurement accuracy, this discrepancy is taken into account during data analysis. It is reported that the most recent da Vinci V robot is already equipped with digital encoders, which have greatly improved positioning accuracy. Nevertheless, addressing this problem remains important for researchers who use the first-generation da Vinci research kit (dVRK) \cite{dvrk_reference_Peter} as a testing platform. Even in the future, when researchers migrate to the dVRK Si, similar positioning inaccuracies are likely to persist. Additionally, recent years have seen an increasing number of similar cable-driven surgical robots enter the market, and hence the development of a generalizable methodology holds greater commercial potential.

\subsection{Related works}
\subsubsection{Classic approach}
The classic way to address the hand-eye calibration problem is to adopt an $\matfont{AX=XB}$ \cite{AX=XB_1987}, or $\matfont{AX=YB}$ \cite{AX=YB_1994} formulation, where $\matfont{X}$, $\matfont{Y}$ are the unknown matrices, and $\matfont{A}$, $\matfont{B}$ are measured by an external sensing device. The overarching goal is to estimate $\matfont{X}$, the hand-eye transformation. Different mathematical representations can be leveraged to simplify the original equation by reducing the number of unknown parameters. These representations include dual-quaternions \cite{handeye-dual-quaternion}, Lie groups \cite{hand-eye-lie-groups}, screw motions \cite{hand-eye-screw-motion}, Kronecker operator \cite{hand-eye-kronecker}, among others, which implicitly determine whether the rotation and translation components are estimated sequentially or simultaneously. 

These hand-eye calibration solvers can be further divided based on whether an analytical or numerical approach is adopted. Zhang \etal \cite{zhang2017computationally} proposed a computationally efficient approach that iteratively estimates the rotation and translation components, represented by dual-quaternions, until a convergence criterion is met. Pachtrachai \etal \cite{pachtrachai2018adjoint} also proposed an algorithm that alternately estimates rotation and translation using twist motions to achieve higher accuracy. Most methods in this category, however, require an external calibration object to provide $\matfont{A}$ and $\matfont{B}$ measurements. These objects, whether directly attached to the robot \cite{ozguner2020camera}, or wrapped around the tool shaft \cite{cartucho2022enhanced}, can be difficult to be adopted in a real clinical setup. To overcome this limitation, recent research has focused on marker-free hand-eye calibration approaches \cite{pachtrachai2021learning, zhou2018towards}, where the instrument and the robot themselves serve as calibration objects, and $\matfont{A}$ and $\matfont{B}$ are calculated through tool localization and pose estimation. Consequently, the accuracy of these methods may degrade significantly when the pose estimation accuracy is poor. 


\subsubsection{Estimator-based approach}
Since encoder readings and forward kinematics already exist for most surgical robots, research has also turned to direct calibration methods via estimators that fuse vision and kinematics data. The advantage of such a hybrid approach is that it provides more consistent pose estimation results, especially in the presence of tool articulations and occlusions, as stated in \cite{Max_Allan}. Most estimator-based methods differ in the selection of statistical estimators and visual features incorporated in the estimation framework. Ye \etal \cite{MengLong_2016_MICCAI} first proposed an Extended Kalman Filter (EKF) framework where key point features obtained via template matching were leveraged to calibrate $\matfont{X}$. Similarly, Moccia \etal \cite{Fanny_Virtual_Fixtures} developed an EKF framework that utilized the 3D instrument tip position reconstructed from stereo image triangulation. Lu \etal \cite{UCSD_keypoints_2022} then proposed a Particle Filter (PF) framework that incorporated key points detected via a pre-trained deep neural network. 

Although key points are arguably the most accessible features on the image plane, most data-driven key points-based methods require labor-intensive human labeling. Richter \etal \cite{UCSD_Unified} further incorporated tool edge features into the PF framework, and Liang \etal \cite{liang2025efficient} demonstrated the importance of edge features. However, obtaining instrument edge features is challenging in real clinical scenarios. First, the endoscopic camera most often focuses closely on the instrument tip part. Second, the most common way to extract edge features is via obtaining instrument masks followed by selecting the longest line using the Canny edge detector \cite{Canny_edge_detection}, which is both time-consuming and noise-sensitive. Furthermore, some segmentation algorithms, such as the Segment Anything Model \cite{SAM_2023}, which was adopted in \cite{liang2025differentiable}, falls short of accuracy for surgical applications \cite{he2023accuracy}.

Additionally, instrument silhouettes can also serve as features to assist pose estimation. Hao \etal \cite{Hao_silhouette_rendering} proposed a PF framework where virtually rendered instrument silhouettes were compared with tool segmentation results to select particles assigned with accurate parameter estimations. They reported that expensive tool-rendering process limited the framework to 0.3 Hz, making it impractical for real-time applications without relying on external acceleration resources.

\subsubsection{Others}
More data-driven approaches have also emerged in recent years, in which a pre-trained model is used to produce end-to-end estimates of \matfont{X} and the joint angles. Liang \etal \cite{liang2025differentiable} proposed a differentiable rendering-based method that iteratively calculates the optimal candidate by aligning rendered instrument silhouettes with reference masks, given a reliable initialization. Yang \etal \cite{yang2025instrument} developed a render-and-compare pose estimation framework that incorporates Gaussian Splatting to cater for a monocular setup. Fan \etal \cite{fan2024reinforcement} proposed a Reinforcement Learning-based method where an agent is trained to align the pixel projections of key points on a virtual instrument with the corresponding observations in the image plane. Although promising results were achieved, due to the specific shape of the instrument, approaches based on silhouette rendering often encounter inaccuracies under poor initializations or when trapped in local minima.

\begin{figure*}[h]
    \centering
    \includegraphics[width=0.85\textwidth]{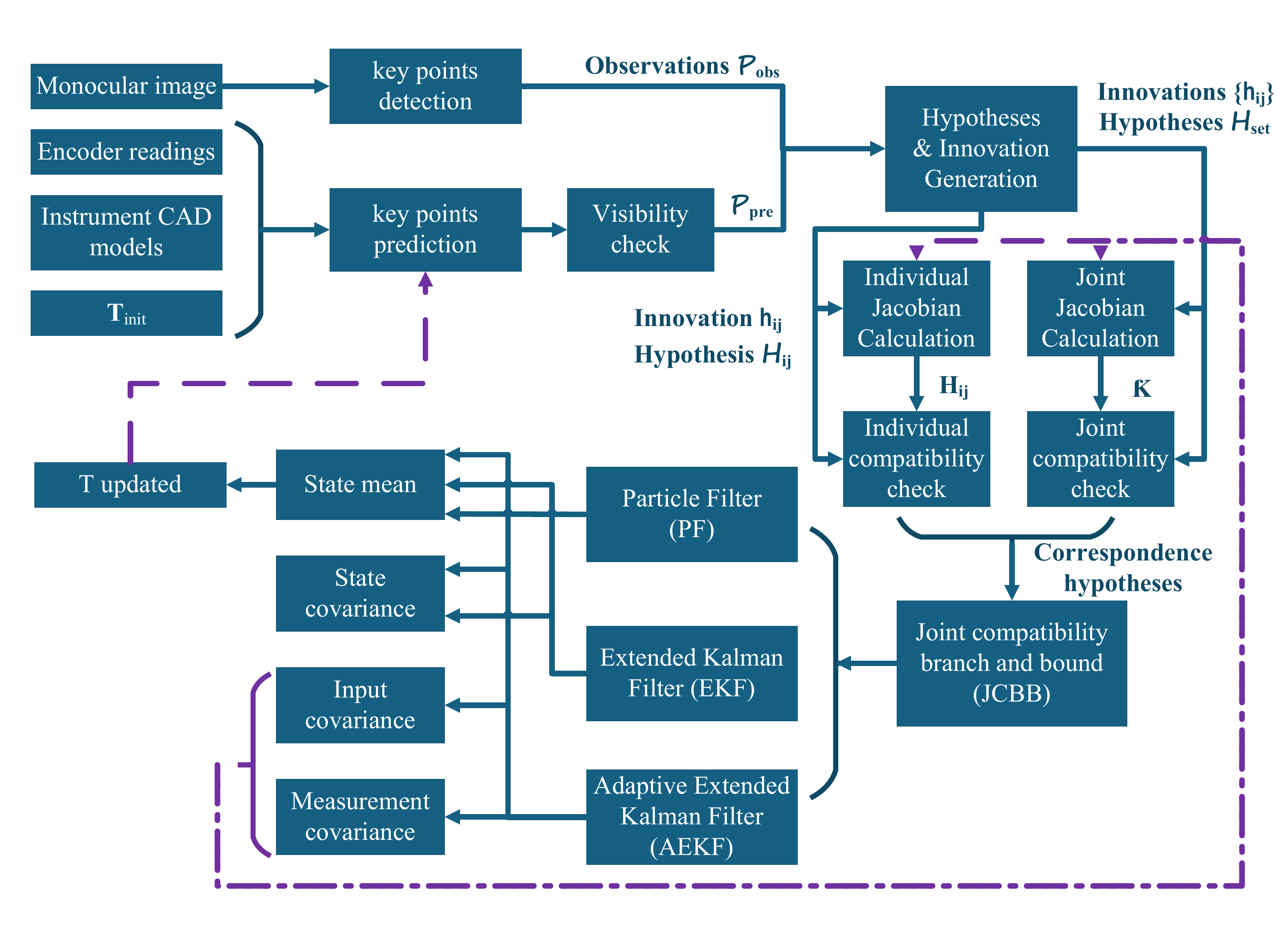}
    \caption{Framework Overview}
    \label{fig:framework overview}
\end{figure*}

%% file: Files/Methods.tex
 \section{Methods} \label{sec:methods}


\begin{figure}[ht]
    \centering
    \includegraphics[height=4.0cm]{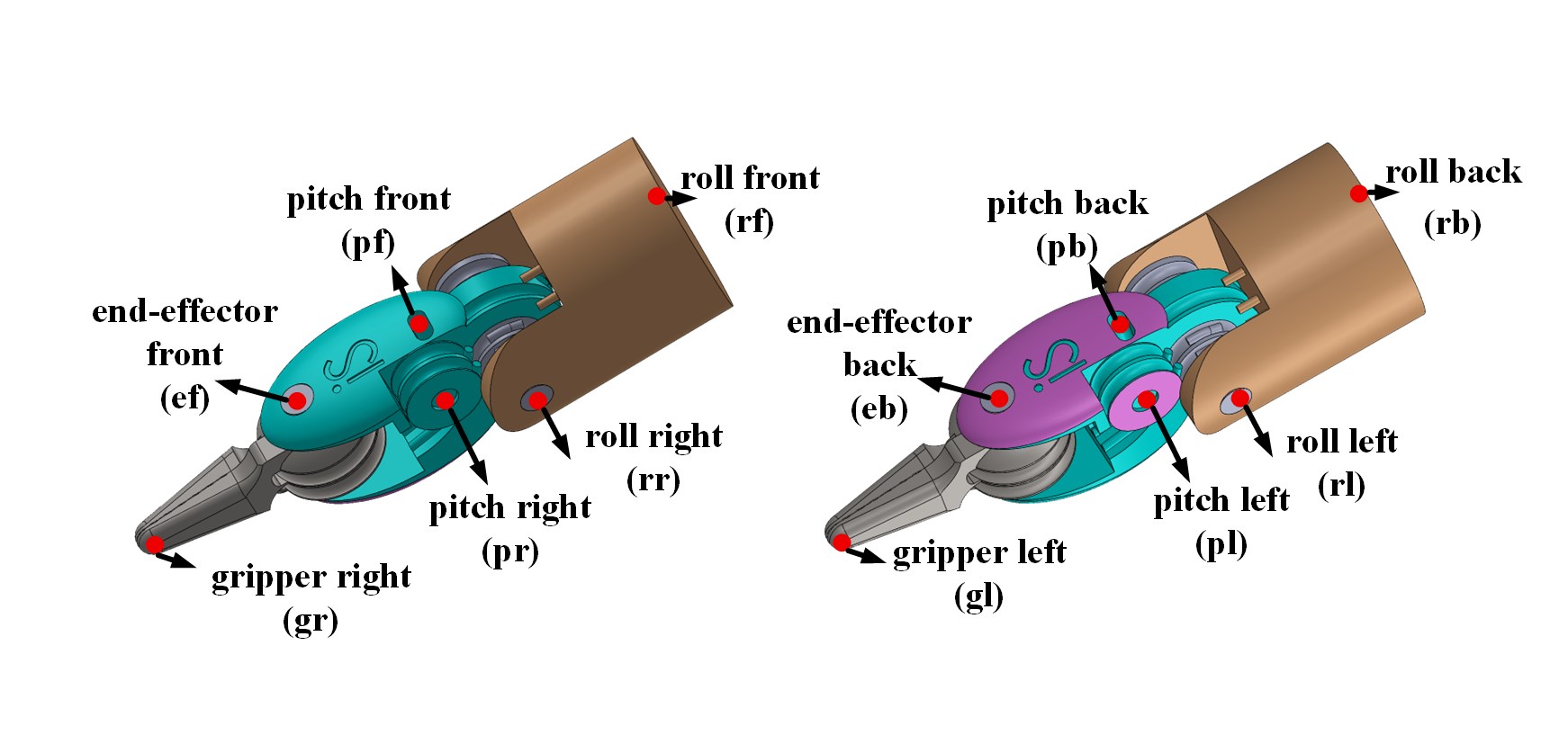}
    \caption{Illustration of key points on the instrument.}
    \label{fig:keypoints}
\end{figure}

\begin{figure}[h]
    \centering
    \includegraphics[width=0.48\textwidth]{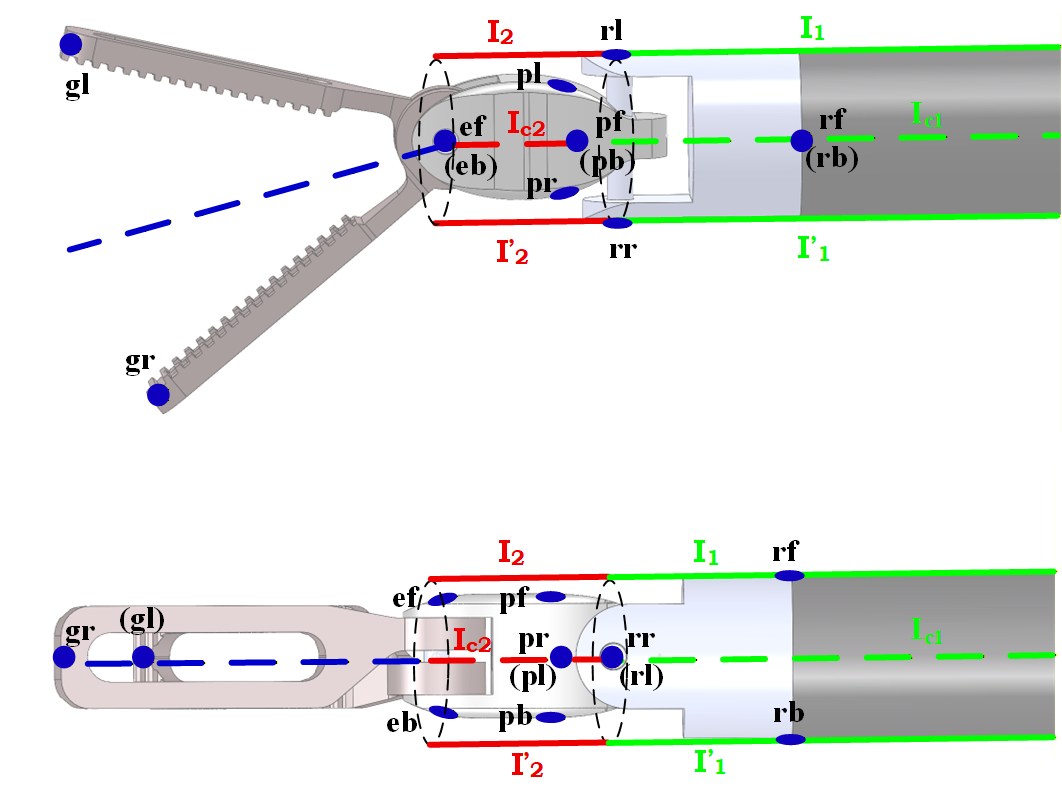}
    \caption{Illustration of key point projection onto the image plane. The top and bottom images represent situations in which the top and right sides of the instrument are directly facing the camera, respectively. $\vecfont{I}_1,\vecfont{I}'_1,\vecfont{I}_{c_1}$ represent the edges and central line of the ``roll-segment" cylinder projection, and $\vecfont{I}_2,\vecfont{I}'_2,\vecfont{I}_{c_2}$ represent those that of the ``pitch-segment" cylinder projection.}
    \label{fig:visibility_check}
\end{figure}

\subsection{Framework overview}
The overall framework is illustrated in Fig.\ref{fig:framework overview}. Before operation, key points are defined on the instrument CAD model, as illustrated in Fig.\ref{fig:keypoints} and Fig.\ref{fig:visibility_check}. During operation, key points in monocular images are detected and processed to be paired with their labels through the visibility check and JCBB algorithm blocks (Subsections. \ref{subsec:JC}, \ref{subsec:JCBB}, and \ref{subsec:VS}). Key points association is illustrated in Fig.\ref{fig:kp associations}. These paired correspondences are subsequently sent to the state estimation block (Subsection. \ref{subsec:StateEstimation}) to obtain the calibrated hand-eye transformation matrix.

\input{Files/Tables/Nomenclature_transformation}

\subsection{Preliminaries}
\subsubsection{Coordinates transformation}
\begin{figure*}[ht]
    \centering
    \includegraphics[width=0.8\textwidth]{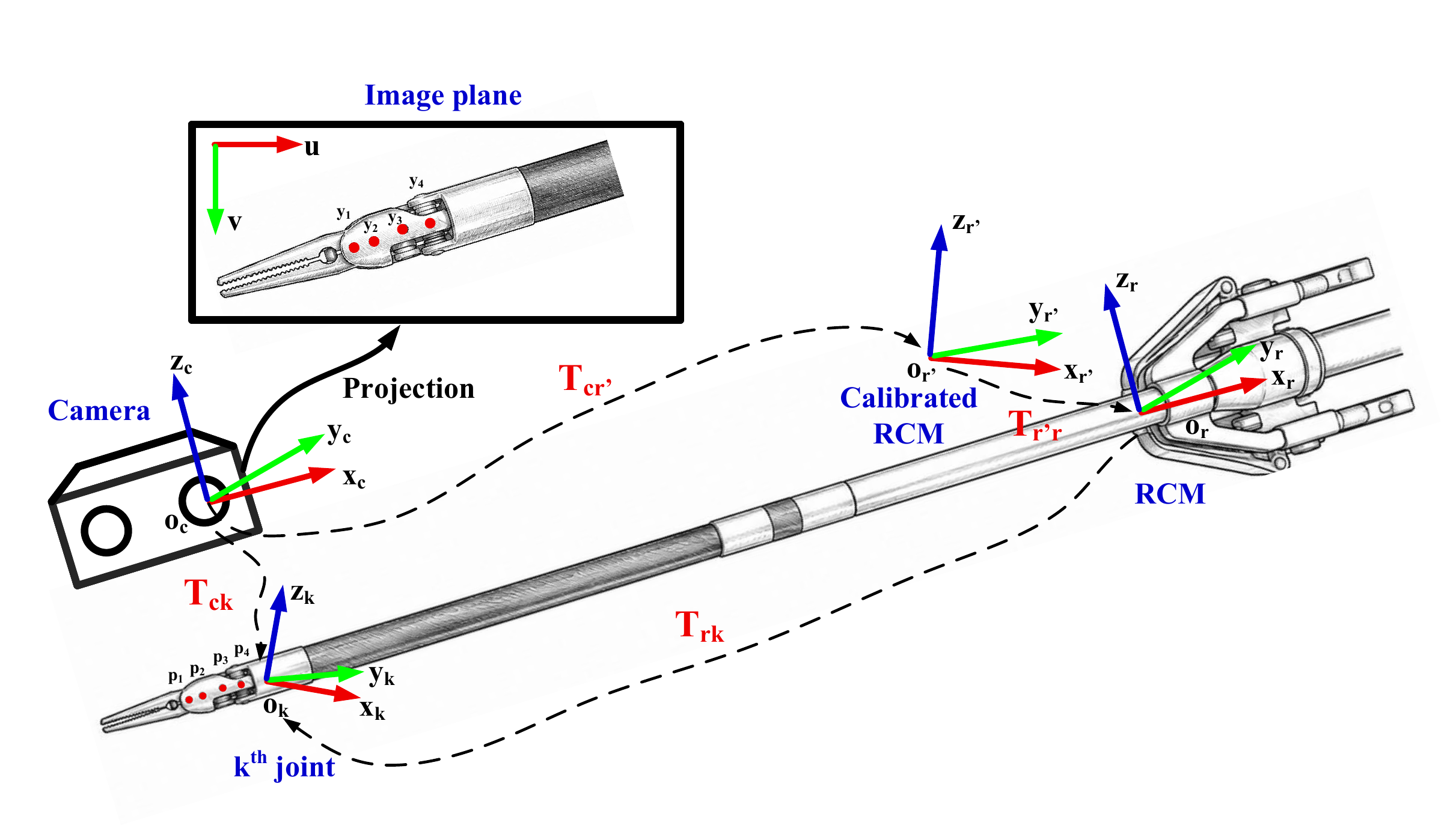}
    \caption{\add{Illustration of frame transformations. Transformation matrices $\matfont{T}_{cr'}$, $\matfont{T}_{r'r}$, $\matfont{T}_{rk}$, and $\matfont{T}_{ck}$ are illustrated. 3D key points $\vecfont{p}_1$,$\vecfont{p}_2$,$\vecfont{p}_3$,$\vecfont{p}_4$ on the instrument are projected onto the image plane as $\vecfont{y}_1$,$\vecfont{y}_2$,$\vecfont{y}_3$, and $\vecfont{y}_4$, respectively.}}
    \label{fig:frame_transform}
\end{figure*}

Altogether, there exist 6 parameters that consist of an additional homogeneous $4\times4$ transformation matrix $\matfont{T}_{r'r}$. $\vecfont{x}_{r'r}=[(\mathbf{\Theta}_{r'r})^\textmd{T}, {(\vecfont{t}_{r'r})}^\textmd{T}]^\textmd{T}$, stores these unknown parameters. $\mathbf{\Theta}_{r'r}=[\alpha, \beta, \gamma]^\textmd{T}$, represents $\textmd{Z-Y-X}$ Euler angles, and $\vecfont{t}_{r'r} = [x_{r'r},y_{r'r},z_{r'r}]^\textmd{T}$, represents translational components. $\matfont{T}_{r'r}$ can be mathematically expressed using $\vecfont{x}_{r'r}$, in eq.\ref{eq:T_and_x}. \add{A notation table for the transformation chain is provided in TABLE.\ref{tab:nomanclature_transformation}. Frame transformations are illustrated in Fig.\ref{fig:frame_transform}.}
\input{Files/Equations/T_and_x}

\begin{figure*}[h]
    \centering
    \includegraphics[width=\textwidth]{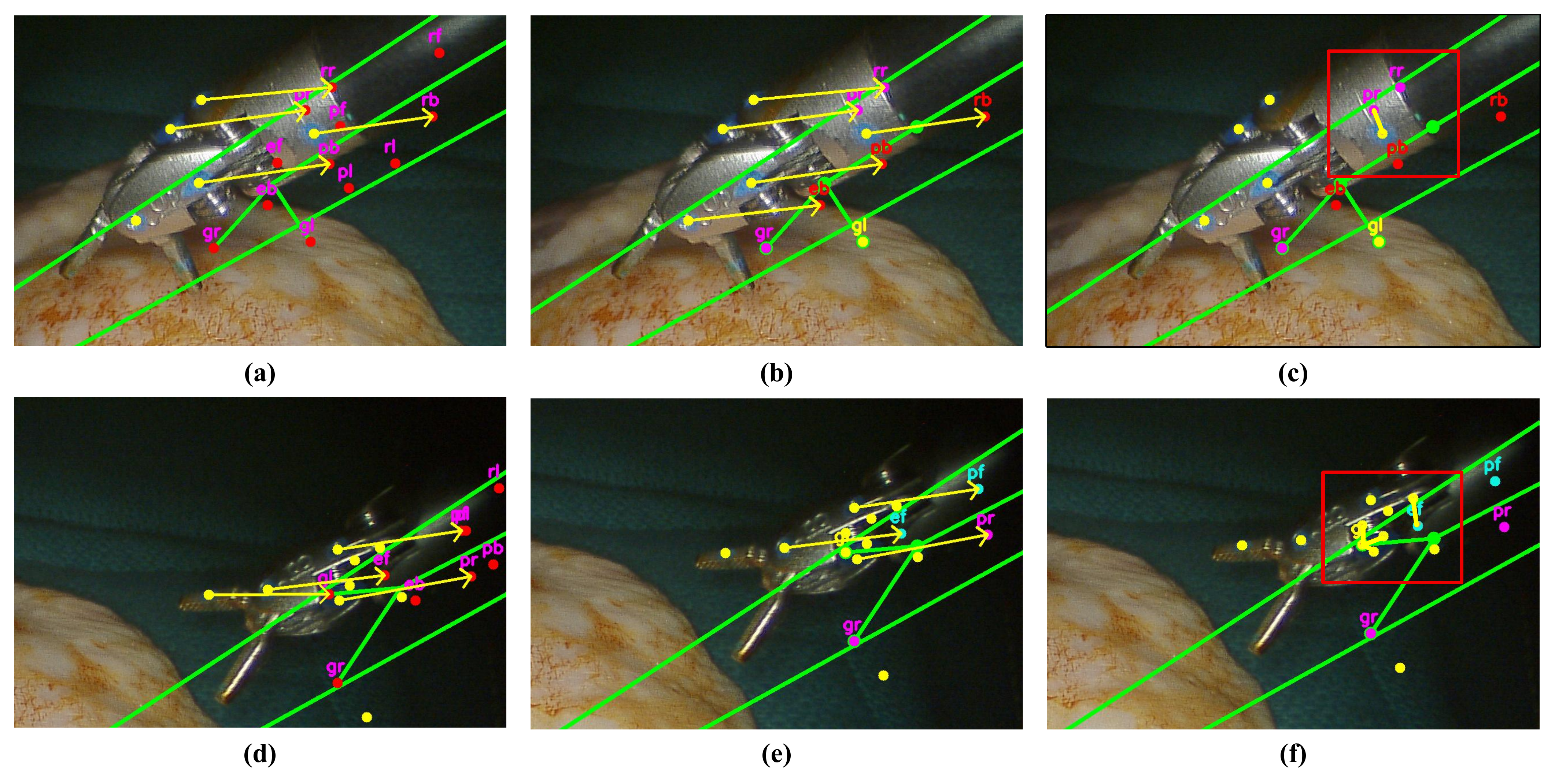}
    \caption{\add{(a),(b),(d),(e) illustrate correct key point association results using the JCBB approach. In (a) and (d), the red points represent key point projections via the hand-eye transformation matrix; the yellow points represent features detected on the image plane including outliers, and the arrows indicate established associations. In (b) and (e), only visible candidate key point projections are presented. (c) and (f) illustrate incorrect association results obtained using the K-Nearest Neighbours method (K=3), as highlighted in the red boxes.}}
    \label{fig:kp associations}
\end{figure*}

Specifically, $\matfont{R}_{r'r}(\alpha,\beta,\gamma)$ can be expanded into three sequential rotations around their individual axes, in eq.\ref{eq:Rrr_expand}, where $\textmd{c}$ and $\textmd{s}$ represent $\textmd{cos}$ and $\textmd{sin}$ operators, respectively. 
\input{Files/Equations/Rot_mat_ZYX} 

The initial hand-eye calibration matrix is represented as $\matfont{T}_{cr'}$, in eq.\ref{eq:T_init}. 
\input{Files/Equations/T_init}

$\vecfont{p}$ is a key point located relative to the $k^{th}$ PSM joint, with its local position $\vecfont{P}_{k} = [x_{k},y_{k},z_{k},1]^\textmd{T}$. $\vecfont{p}_c=[(\vecfont{t}_c)^\textmd{T},1]^\textmd{T}=[x_c,y_c,z_c,1]^\textmd{T}$, and $\vecfont{p}_r=[(\vecfont{t}_r)^\textmd{T},1]^\textmd{T}=[x_r,y_r,z_r,1]^\textmd{T}$, represent its homogeneous 3D positions in camera frame and PSM base frame, respectively. Since transformation $\matfont{T}_{rk}$ can be directly calculated from PSM encoder readings and Denavit-Hartenberg(DH) parameters, we have eq.\ref{eq:pc}.

\input{Files/Equations/pc_1}

In reality, it is the pixel projection of $\vecfont{p}_c$ on the image plane that serves as observation, represented as $\vecfont{y}=[u,v]^\textmd{T}$. When no camera distortions are present, we have eq.\ref{eq:camera_bp}.

\input{Files/Equations/BP_camera}

\subsubsection{Jacobian calculations}
Having established the relationship between visual measurements and unknown parameter inputs, here we derive essential Jacobian matrices that are required by subsequent procedures, from eq.\ref{eq:pc} and \ref{eq:camera_bp}.

We first derive the Jacobian matrix $\matfont{H}_{robot}$, which operates between $\vecfont{p}_c$ and $\vecfont{x}_{r'r}$. $\matfont{H}_{robot}$ consists of partial differentiations over both rotational component $\mathbf{\Theta}_{r'r}$ and translational component $\vecfont{t}_{r'r}$, summarized in
eq.\ref{eq:Jacobian_part1}. Since the relationship between $\vecfont{p}_c$ and $\vecfont{t}_{r'r}$ is linear, we have eq.\ref{eq:Jacobian_part2}.
\input{Files/Equations/Jacobian_part1}

Subsequently, we derive each column of $\matfont{H}_\mathbf{{\Theta}}$ based on linear algebra through eq.\ref{eq:Jtheta_1} to \ref{eq:Jtheta_4}.
\input{Files/Equations/Jacobian_part2}

Then we derive the the intermediary Jacobian matrix $\matfont{H}_{cam}$, which operates between observation $\vecfont{y}$ and $\vecfont{p}_c$. Taking derivatives from eq.\ref{eq:camera_bp}, we have eq.\ref{eq:Jacobian_obs_c}.

\input{Files/Equations/Jacobian_part3}

Finally, we have the overall Jacobian matrix $\matfont{H}_{2\times6}$, which operates between observation \add{$\vecfont{y}$} and $\vecfont{x}_{r'r}$, in eq.\ref{eq:Jacobian_final}.

\input{Files/Equations/Jacobian_part4}

To summarize, for each key point $\vecfont{p}$, as long as we have the current \add{state} estimation $\vecfont{x}_{r'r}$, its pixel observation $\vecfont{y}$, and its 3D position in PSM base frame $\vecfont{p}_r$, we can calculate its Jacobian matrix $\matfont{H}(\vecfont{x}_{r'r}, \vecfont{y})$, which is essential to the following key points association procedure. To simplify annotations, we use $\vecfont{x}$ for $\vecfont{x}_{r'r}$, \add{and $\matfont{H}$ for $\matfont{H}(\vecfont{x}_{r'r}, \vecfont{y})$,}  for the rest of the paper.

\subsection{\add{Compatibility tests}}\label{subsec:JC}
\input{Files/Tables/Nomenclature_JCBB}

\add{In this subsection, the individual and joint compatibility tests are presented, laying the mathematical foundation for subsection.\ref{subsec:JCBB}. Neira \etal \cite{neira2001data_rebuttal} originally proposed this approach to addressing the data association problem in simultaneous vehicle localization and map building. Here we adopt the framework and leverage the analytical Jacobian derived in the previous subsection. A joint compatibility-based approach is preferred over conventional approaches, such as the K-Nearest Neighbours \cite{knn_1967}, because these approaches that rely solely on pixel-level Euclidean distance become less reliable in the presence of closely-located symmetrical candidate points and noisy measurements, as shown in Fig.\ref{fig:kp associations}. A notation table for subsection.\ref{subsec:JC} and \ref{subsec:JCBB} is provided in TABLE.\ref{tab:nomanclature_JCBB}.}

\subsubsection{Individual Compatibility test}
Under the Markov assumption \cite{howard2012dynamic}, unknown time-variant parameter $\vecfont{x}$ evolves according to eq.\eqref{eq:monte-model1}, where $p$ represents state forward function, $\vecfont{u}$ represents current input, and $\vecfont{e}$ represents Gaussian input noises. \add{In this study, no current input is expected.} In eq.\eqref{eq:monte-model2}, $\vecfont{y}$ represents current measurement, $g$ represents mapping function, and $\vecfont{v}$ represents Gaussian measurement noises.

\input{Files/Equations/JC}

\add{At time $t$, there are $m$ key point observations in the measurement list. ${\mathcal{P}_{obs}}=\{[\hat{u}_1,\hat{v}_1]^\textmd{T}...[\hat{u}_m,\hat{v}_m]^\textmd{T}\}$, and $n$ predicted key point features in the prediction list ${\mathcal{P}_{pre}}=\{[\check{u}_1, \check{v}_1]^\textmd{T}...[\check{u}_n,\check{v}_n]^\textmd{T} \}$.$\mathcal{H}_{ij}$ represents the hypothesis that the $i^\text{th}$ observation, $\mathcal{P}_{obs,i}$, is associated with the $j^\text{th}$ prediction, ${\mathcal{P}_{pre,j}}$. $\mathcal{H}_{i0}$ indicates that no key point prediction matches, implying that the $i^{th}$ observation is an outlier and that $\mathcal{H}_{i0}$ is a trivial hypothesis. We now evaluate the likelihood of $\mathcal{H}_{ij}$ via individual compatibility test. The innovation for $\mathcal{H}_{ij}$ is defined as $\vecfont{h}_{ij}=\mathcal{P}_{obs,i} - \mathcal{P}_{pre,j}$. We define $\check{\vecfont{x}}$ as the current state estimation, and $\bar{\vecfont{x}}$ as the ground truth of $\check{\vecfont{x}}$. Similarly, we define $\hat{\vecfont{y}}$ as current observation, and $\bar{\vecfont{y}}$ as the ground truth.\footnote{\add{hat $\hat{}$ indicates measurement values (a posteriori); bar $\bar{}$ indicates ground truth; check operator $\check{}$ indicates prediction values (a priori); superscript $^{(t)}$ indicates time instance.}}}

We then define an implicit measurement function \add{$f(\bar{\vecfont{x}},\bar{\vecfont{y}})=\bar{\vecfont{y}}-g(\bar{\vecfont{x}})=\vecfont{0}$}, for hypothesis $\mathcal{H}_{ij}$. After applying the first-order Taylor expansion and conducting linearization at \add{$(\check{\vecfont{x}}, \hat{\vecfont{y}})$}, we have eq.\eqref{eq:implicity_measure}.

\input{Files/Equations/ImplicitMeasure}

$\matfont{H}_{ij}$ represents the Jacobian matrix that corresponds to hypothesis $\mathcal{H}_{ij}$, evaluated at $\add{\check{\vecfont{x}}}$. According to the previous subsection, the value of \add{$\matfont{H}_{ij}$ } also depends on the 3D position of the $i^{th}$ labeled key point in PSM frame as well as pixel values of the $j^{th}$ measurement, hence is directly affected by the verity of $\mathcal{H}_{ij}$. We use $\matfont{C}_{ij}$ to represent the covariance matrix for innovation $\vecfont{h}_{ij}$ in eq.\eqref{eq:C_mat_indi}.

\input{Files/Equations/Cij_individual}

To evaluate the credibility of $\mathcal{H}_{ij}$, we introduce squared Mahalanobis distance as eq.\eqref{eq:Mahalanobis}, where $d=\text{dim}(\vecfont{y})$ and $\alpha$ is threshold confidence level. If the calculated Mahalanobis distance falls within the confidence interval of a chi-square distribution, the presumed hypothesis is deemed credible.

The advantage of adopting Mahalanobis distance as evaluation criterion over simple Euclidean distance is that they provide a better alignment description in statistical space \cite{mclachlan1999mahalanobis}. This advantage becomes more evident when it comes to multiple pairings.

\subsubsection{Joint Compatibility test} 
The previous subsection focuses on evaluating individual compatibility for a single association hypothesis. However, in reality, there usually exists multiple unlabeled observations with outliers. Moreover, being individually compatible does not necessarily lead to a globally compatible outcome. To overcome these limitations, in this subsection, we expand the previous individual compatibility evaluation criterion to cases where multiple associations are present. The algorithm for conducting joint compatibility test is presented as Algorithm.\ref{alg:JC_algorithm}.

We assume that there exists \add{joint hypotheses} \add{$\Hcand$}, which contains $k$ non-trivial association hypotheses. These non-trivial hypotheses are denoted as $\{\mathcal{H}_{i_1j_1}...\mathcal{H}_{i_kj_k}\}$. For each non-trivial \add{individual} hypothesis, we first calculate its corresponding innovation and Jacobian matrix and then stack them vertically. Their corresponding $\mathbf{\Sigma_e}$ and $\mathbf{\Sigma_v}$ matrices are stacked diagonally. Then the covariance matrix $\matfont{C}_{\Hcand}$ for all innovations is calculated as per eq.\eqref{eq:C_mat_joint}.

\input{Files/Equations/Cij_joint}

The squared Mahalanobis distance $D^2_{\Hcand}$ for $\Hcand$ is expressed in eq.\eqref{eq:Mahalanobis_joint}.

\input{Files/Equations/Mahalanobis_joint}

Similar to eq.\eqref{eq:Mahalanobis}, if $D^2_{\Hcand} < \chi^2_{d,\alpha}$, where $d=2k$, $\Hcand$ is deemed jointly compatible. 

Distance $l_{\Hcand}$ is calculated as per eq.\eqref{eq:ML}, representing the negative logarithm of matching likelihood \cite{JCBB_ML}, which was shown as the optimal statistic to be optimized compared with the squared Mahalanobis distance. A smaller $l_{\Hcand}$ indicates a more compact joint association.

\input{Files/Equations/ML}

\subsection{Joint Compatibility Branch and Bound (JCBB)}\label{subsec:JCBB}
In the previous subsection, it is assumed that hypothesis list $\Hcand$ is readily available. However, in reality, to obtain a reliable hypothesis list is challenging. In this subsection, we present an algorithm for obtaining the optimal $\Hcand$ through recursively establishing new hypotheses and self-updating, named Joint Compatibility Branch and Bound (JCBB).

\add{For $\mathcal{P}_{obs,i}$, we initially establish a list of $n$ individual hypotheses denoted as $\Hcand_i=\{ \mathcal{H}_{i1}... \mathcal{H}_{in}\}$, and altogether, we have $m \times n$ individual hypotheses for all $m$ observations, denoted as $\Hlist=\{ \Hcand_1 ...\Hcand_m \}$. After conducting individual compatibility check, as per eq.\eqref{eq:Mahalanobis}, spurious hypotheses are pruned and outlier observations are determined. $\Hlist$ now reduces to $\{\{\mathcal{H}_{11}...\mathcal{H}_{1j_1}$\}...\{$\mathcal{H}_{m1}...\mathcal{H}_{mj_m}\}$\}, where $\{j_1...j_m\} $ are indices smaller than $n$.}

The JCBB algorithm reads input as an empty list \add{$\Hcand=\{\}$}, and returns the final list of associations, \add{$\Hfinal$}, which possesses the highest number of non-trivial associations, $n_{pair}$, and the smallest negative logarithm of matching likelihood \add{$l_{\Hcand}$}. The process alternates between adding new hypothesis to $\Hcand$, and conducting prompt joint compatibility check, as per Algorithm.\eqref{alg:JC_algorithm}, until the size of $\Hcand$ reaches $m$. Algorithm \ref{alg:JCBB_algorithm} summarizes how $\Hcand$ accumulates and self-updates in a recursive manner. \add{To facilitate algorithm implementation, in line \ref{line:JCBB_pair_max_pair_1} and \ref{line:JCBB_pair_max_pair_2}, the maximum number of remaining pairs is calculated as a baseline, since each observation cannot be assigned to more than one labels.}

\input{Files/Algorithms/JC_evaluation}
\input{Files/Algorithms/JCBB_evaluation}

\subsection{Key points visibility test}\label{subsec:VS}
The computational cost of the JCBB algorithm implementation depends on the size of $\mathcal{P}_{pre}$ and $\mathcal{P}_{obs}$. This subsection introduces an additional visibility-check block for facilitating the key point association process by reducing the size of $\mathcal{P}_{pre}$, based on the observation that key points located on four different sides of an instrument cannot be all visible simultaneously. 

As illustrated in Fig.\ref{fig:keypoints}, twelve key points are labeled on the wrist section of a cylindrical instrument and can be classified by their position on the instrument, namely, front, back, left, and right. Intuitively, when key points located on the front (or left) side are directly exposed to the camera view, those on the back (or right) side are not visible, except in rare cases where the instrument rotates to a specific pose such that key points on two opposite sides align with two edge boundaries, making three sides visible simultaneously. In most cases, however, at most two sides are visible in the camera view. 


The first step is to determine the number of visible sides in the current camera view based on a central-edge distance ratio check. We use $\vecfont{I}, \vecfont{I}'$, and $\vecfont{I}_{c}$ to represent the upper edge, lower edge, and center line of the cylindrical instrument on an image plane, respectively. The calculation of $\vecfont{I}$ and $\vecfont{I}'$ were presented in \cite{lu2022unified}, and $\vecfont{I}_{c}$ is the line that connects the projection of certain instrument joints on the image plane. $\mathcal{O}_{roll}=\{\mathbf{o_{rf}}, \mathbf{o_{rb}, \mathbf{o_{rl}}, \mathbf{o_{rr}}} \}$ represents the pixel projection of all key points from the ``roll" family. We also have $\mathcal{O}_{pitch}=\{\mathbf{o_{pf}}, \mathbf{o_{pb}, \mathbf{o_{pl}}, \mathbf{o_{pr}}} \}$, $\mathcal{O}_{end}=\{\mathbf{o_{ef}}, \mathbf{o_{eb}} \}$  and $\mathcal{O}_{grip}=\{\mathbf{o_{gl}}, \mathbf{o_{gr}} \}$, representing ``pitch", ``end-effector" and ``gripper" families, respectively. 

For example, \add{$d_{rf,c}$} represents the center distance between $\mathbf{o_{rf}}$ and $\vecfont{I}_{c_1}$, and \add{$d_{{rf},edge} = min ( d_{{rf},{\vecfont{L}_1}}, d_{{rf},{\vecfont{L}_2}})$}, represents the closest edge distance between $\mathbf{o_{rf}}$ and two edge boundaries. Similar notations also apply to the other key points. The center-edge ratio $\eta_{fb}$, for front-back sides, and $\eta_{lr}$, left-right sides, are defined in eq.\eqref{eq:eta_fb} and \eqref{eq:eta_lr}, respectively. $\gamma$ is a predefined ratio threshold. If either $\eta_{fb} \geq \gamma$ or $\eta_{lr} \geq \gamma$, it indicates that one side is predominantly visible; otherwise two sides are visible. 

\input{Files/Equations/visibility_check}

\begin{figure*}[h]
    \centering
    \includegraphics[width=0.85\textwidth]{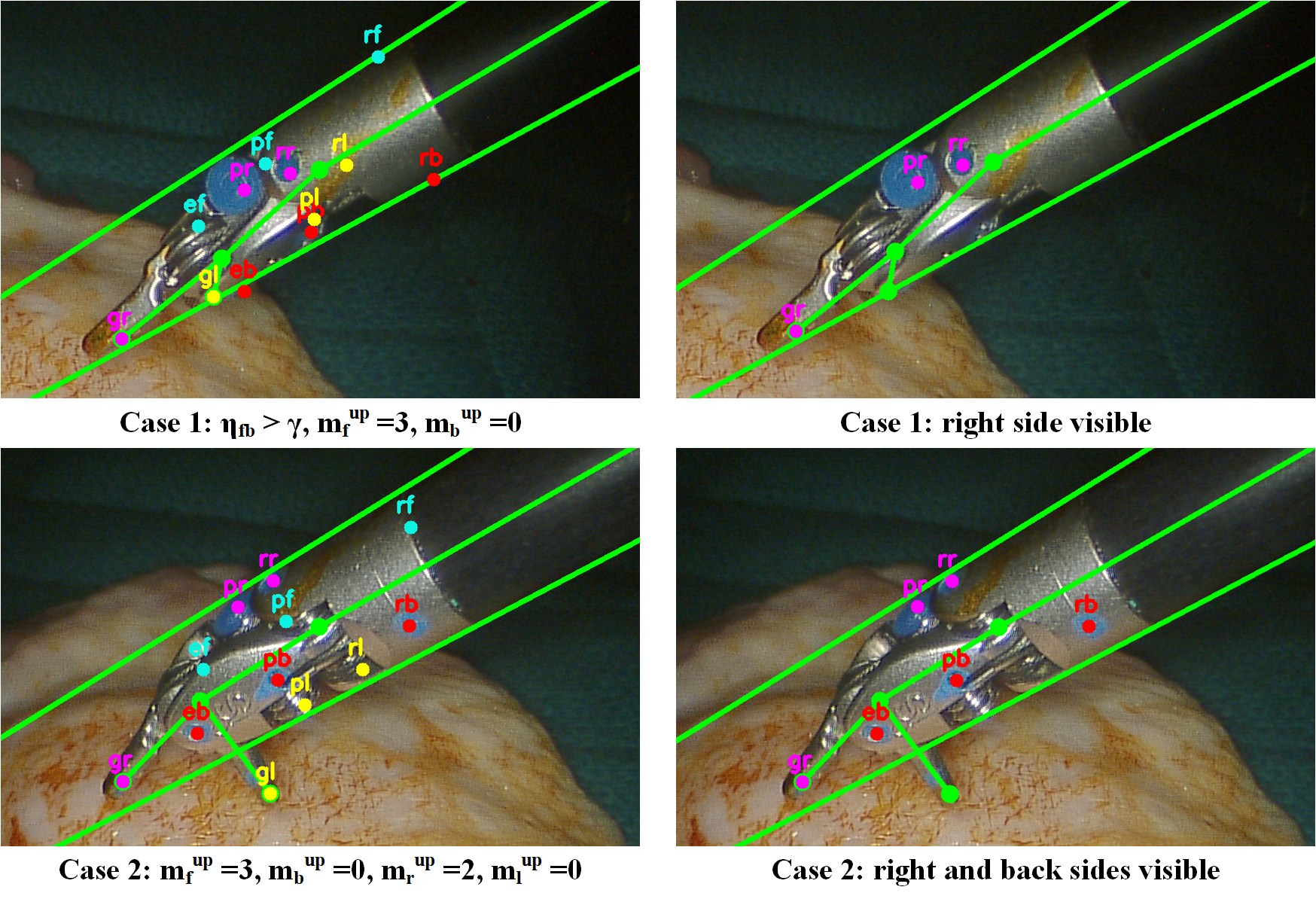}
    \caption{Visibility check examples. After conducting visibility check, the number of candidate key points is reduced.}
    \label{fig:visibility_check_pruning}
\end{figure*}

\begin{figure*}[h]\label{fig:decision_tree}
    \centering
    \includegraphics[height=7cm]{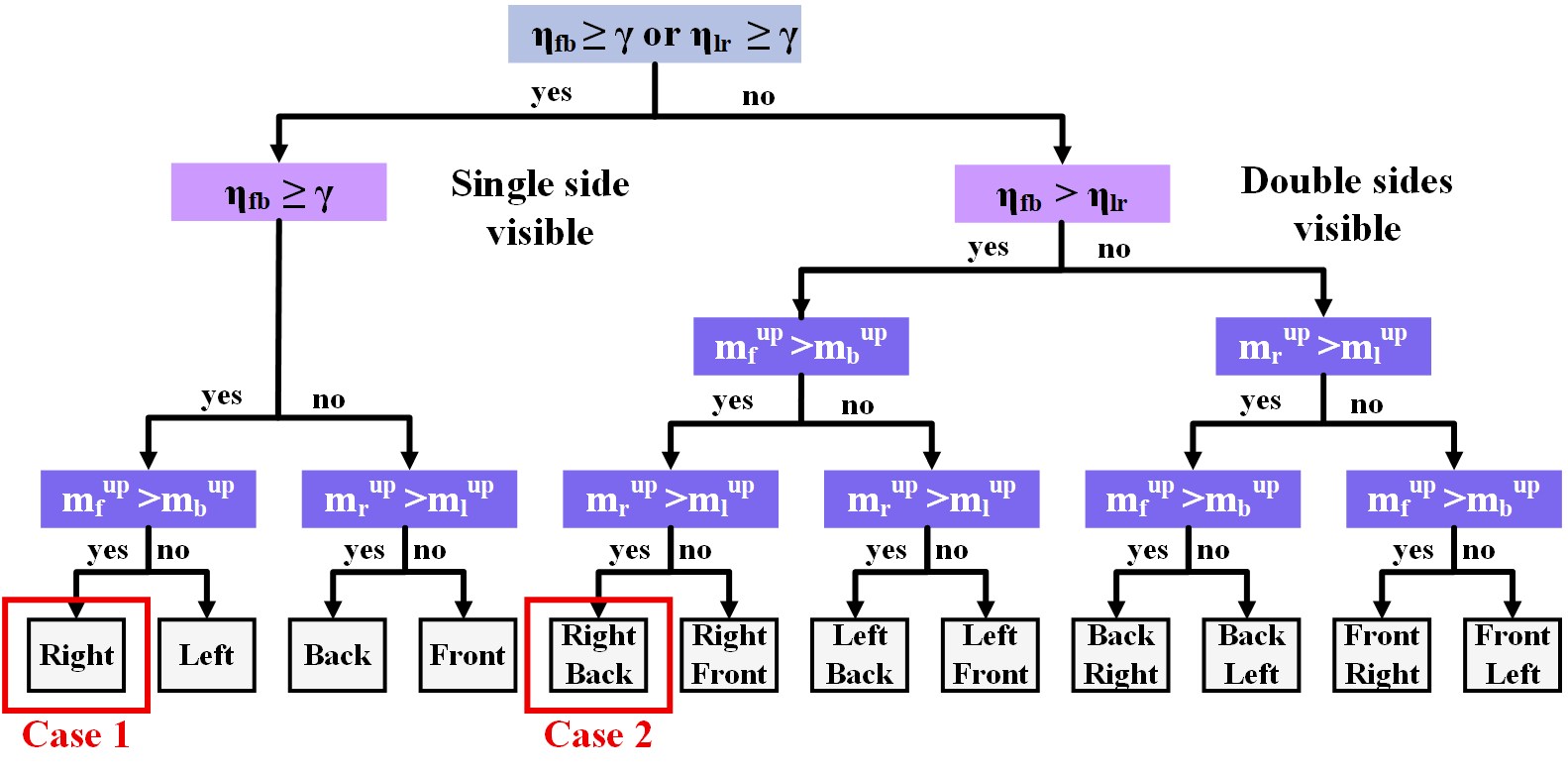}
    \caption{Visibility check decision tree}
    \label{fig:VC_decision_tree}
\end{figure*}

The second step is to determine which sides are visible by comparing the distribution of key points in relation their corresponding central line and edge boundaries. For key points in $\mathcal{O}_{roll}$, the central line is $\vecfont{I}_{c_1}$, and the edge boundaries are $\vecfont{I}_{1}$ and $\vecfont{I}'_{1}$; for key points in $\mathcal{O}_{pitch}$ and $\mathcal{O}_{end}$, the central line is $\vecfont{I}_{c_2}$, and the edge boundaries are $\vecfont{I}_{2}$ and $\vecfont{I}'_{2}$, as illustrated in Fig.\ref{fig:visibility_check}. For example, we denote $m_f^{up}$ as the number of front-side key points in $\{\mathcal{O}_{roll}, \mathcal{O}_{pitch}\}$, that are either positioned between $\vecfont{I}_{c_1}$ and $\vecfont{I}_1$, or between $\vecfont{I}_{c_2}$ and $\vecfont{I}_2$. Similar notations also apply to the other key points. If it is determined that $\eta_{fb}\geq\gamma$ from step 1, we then compare $m_f^{up}$ with $m_b^{up}$ to determine whether it is the right or left side that is dominantly visible. In cases where two sides are visible, comparisons between $m_f^{up}$ and $m_b^{up}$, as well as $m_l^{up}$ and $m_r^{up}$, are required. An illustration of pruning invisible key points is shown in Fig \ref{fig:visibility_check_pruning}, and the decision tree for conducting visibility check is shown in Fig.\ref{fig:VC_decision_tree}. Eventually, key points on invisible sides are removed from $\mathcal{P}_{pre}$.


\subsection{State estimation with known correspondences}\label{subsec:StateEstimation}
Having established known correspondences, in this subsection, three statistical estimator-based approaches are presented and compared, for estimating the state vector in different realistic scenarios. \add{A notation table for the state update module is provided in TABLE.\ref{tab:nomanclature_filter}.}

\input{Files/Tables/Nomenclature_filter}

\subsubsection{Extended Kalman Filter}
After data associations, \add{we have established the final joint hypotheses $\Hfinal$.} The implementation of Extended Kalman Filter (EKF) for state estimation is presented in Algorithm.\ref{alg:EKF_algorithm}. The state transition model is considered linear, and linearization of the measurement model has already been introduced, as per eq.\eqref{eq:Jacobian_part1} to \eqref{eq:Jacobian_final}. EKF assumes that only Gaussian noises exist during state transition and measurements collection and that $\mathbf{\Sigma_e}$ and $\mathbf{\Sigma_v}$ remain static throughout the process. In \cite{UCSD_Unified}, it was proved that errors in robot joint positions can be lumped into modifications to the current state vector $\xnowprior$, which is continuously estimated, and hence it can be justified that the state transition model is not affected by noisy inputs from the encoder. Eventually, EKF returns the posterior state estimation $\xnow$, state covariance matrix $\sigxnow$, together with innovation list $\mathcal{P}_{inn}$. 

\input{Files/Algorithms/EKF}

\subsubsection{Adaptive Extended Kalman Filter}
The implementation of Adaptive Extended Kalman Filter (AEKF) builds on EKF, as detailed in Algorithm.\ref{alg:EKF_algorithm}. Compared with EKF, AEKF further relaxes the assumption that only Gaussian noises are present, by adaptively updating $\sigenow$ and $\sigvnow$ based on innovations and residuals, respectively. It allows for situations where the camera is suddenly repositioned, or observations contain varying levels of noises.   

\input{Files/Algorithms/AEKF}

\subsubsection{Particle Filter}
The implementation of Particle Filter (PF) for state estimation is presented in Algorithm.\ref{alg:PF_algorithm}. Compared with AEKF, PF further reduces the procedure of measurement model linearization and usually demonstrates stronger resilience in the face of non-Gaussian noises. This is achieved through consistent particle weight updates and particle resampling. However, with an increase in the number of particles selected, the computational cost rises sharply \cite{KLD-ParticleFilter}. Moreover, it sometimes incurs particle degeneration problem \cite{probabilistic_robotics}, where only a handful of particles are assigned with infinitely high weights, representing the actual state, with the rest particles assigned with near-zero weights. In this case, PF fails to provide an accurate state estimation. 

\input{Files/Algorithms/PF}

%% file: Files/Tables/Nomenclature_transformation.tex
\begin{table}[htbp]
\caption{Notation table for the transformation chain}
\centering
\add{
\begin{tabular}{p{1.3cm} p{2.8cm} p{3.4cm}}
\toprule
\textbf{Symbol} & \textbf{Meaning} & \textbf{Frame} \\
\midrule
$\vecfont{p}_c$ & 3D point & camera \\
$\vecfont{p}_r$ & 3D point & uncalibrated PSM base frame  \\
$\vecfont{p}_{r'}$ & 3D point & calibrated PSM base frame  \\
$\vecfont{p}_k$ & 3D point & the $k^{th}$ joint of the robot arm\\
$\matfont{T}_{r'r}$ &transformation matrix & $r'$ to $r$ \\
$\matfont{T}_{cr'}$ &transformation matrix & camera to $r'$ \\
$\matfont{R}_{r'r}$ &rotation matrix & $r'$ to $r$ \\
$\matfont{R}_{cr'}$ &rotation matrix & camera to $r'$ \\
$\vecfont{t}_{r'r}$ &translation vector & $r'$ to $r$ \\
$\mathbf{\Theta}_{r'r}$ &rotation angle vector & $r'$ to $r$ \\
$\vecfont{x}_{r'r}$ & $6 \times 1$ state vector that makes up $\matfont{T}_{r'r}$ & $r'$ to $r$ \\
$\matfont{H}_{robot}$ & $3\times 6$ Jacobian & N/A \\
$\matfont{H}_{cam}$ & $2 \times 3$ Jacobian & N/A \\
$\matfont{H}_{{\ThetaVec}}$ & Jacobian w.r.t rotation & N/A \\
$\matfont{H}_{{\tVec}}$ & Jacobian w.r.t translation & N/A \\
$\matfont{H}$ & overall $2\times 6$ Jacobian & N/A \\
$\matfont{rot}_{x}$ & rotation along the x axis & N/A \\
$\vecfont{y=[u,v]}^\textmd{T}$ & pixel projection & image plane \\ 
\bottomrule
\end{tabular}
}
\label{tab:nomanclature_transformation}
\end{table}

%% file: Files/Equations/T_and_x.tex
\begin{equation}\label{eq:T_and_x}
    \matfont{T}_{r'r} = 
    \begin{bmatrix}
        \matfont{R}_{r'r}(\alpha,\beta,\gamma) & \vecfont{t}_{r'r} \\
        \vecfont{0}_{1\times3} & 1
    \end{bmatrix}
\end{equation}

%% file: Files/Equations/Rot_mat_ZYX.tex
\begin{equation}\label{eq:Rrr_expand}
    \begin{split}
    & \matfont{R}_{r'r}(\alpha, \beta, \gamma) = \matfont{rot}_z(\alpha) \matfont{rot}_y(\beta) \matfont{rot}_x(\gamma) = \\
    &\begin{bmatrix}
        \textmd{c} \alpha \textmd{c} \beta & \textmd{c} \alpha \textmd{s} \beta \textmd{s} \gamma - \textmd{s} \alpha \textmd{c} \gamma & \textmd{c} \alpha \textmd{s} \beta \textmd{c} \gamma + \textmd{s} \alpha \textmd{s} \gamma \\
        \textmd{s}\alpha \textmd{c}\beta & \textmd{s}\alpha \textmd{s} \beta \textmd{s} \gamma+ \textmd{c} \alpha \textmd{c} \gamma & \textmd{s} \alpha \textmd{s} \beta \textmd{c} \gamma - \textmd{c}\alpha \textmd{s} \gamma \\
        -\textmd{s}\beta & \textmd{c}\beta \textmd{s}\gamma & \textmd{c}\beta \textmd{c}\gamma
    \end{bmatrix}
    \end{split}
\end{equation}

%% file: Files/Equations/T_init.tex
\begin{equation}\label{eq:T_init}
    \matfont{T}_{cr'}=
    \begin{bmatrix}
        \matfont{R}_{cr'} & \vecfont{t}_{cr'} \\
        \vecfont{0}_{1\times3} & 1
    \end{bmatrix}
\end{equation}

%% file: Files/Equations/pc_1.tex
\begin{equation}\label{eq:pc}
\begin{split}
    \vecfont{P}_{c} & =\matfont{T}_{cr'} \matfont{T}_{r'r} \matfont{T}_{rk} \vecfont{P}_{k}
    =\matfont{T}_{cr'} \matfont{T}_{r'r} \vecfont{P}_{r} \\
    & = 
    \begin{bmatrix}
        \matfont{R}_{cr'} & \vecfont{t}_{cr'} \\
        \vecfont{0}_{1\times3} & 1
    \end{bmatrix}
    \begin{bmatrix}
        \matfont{R}_{r'r}(\alpha,\beta,\gamma) &  \vecfont{t}_{r'r} \\
        \vecfont{0}_{1\times3} & 1\\
    \end{bmatrix}
    \vecfont{P}_r \\
    & = 
    \begin{bmatrix}
        \matfont{R}_{cr'} \matfont{R}_{r'r}(\alpha,\beta,\gamma) \vecfont{t}_r + \matfont{R}_{cr'} \vecfont{t}_{r'r}
        + \vecfont{t}_{cr'}
        \\
        1 
    \end{bmatrix}
\end{split}
\end{equation}




%% file: Files/Equations/BP_camera.tex
\begin{equation}\label{eq:camera_bp}
\begin{bmatrix}
    u \\
    v \\
    1
\end{bmatrix} = 
\begin{bmatrix}
    f_x & 0 & c_x \\
    0 & f_y & c_y \\
    0 & 0 & 1
\end{bmatrix}
\begin{bmatrix}
    \frac{x_{c}}{z_{c}} \\
    \frac{y_{c}}{z_{c}} \\
    1
\end{bmatrix}
\end{equation}

%% file: Files/Equations/Jacobian_part1.tex

\begin{equation}\label{eq:Jacobian_part1}
    \matfont{H}_{robot} = 
    \begin{bmatrix}
    \matfont{H}_{{\ThetaVec}}, \matfont{H}_{{\tVec}}
    \end{bmatrix}_{3\times6},
    [\matfont{H}_{{\ThetaVec}}]_{ij} = \pd{[\vecfont{P}_c]_i}{[\ThetaVec]_j},
    [\matfont{H}_{{\tVec}}]_{ij}= \pd{[\vecfont{P}_c]_i}{[\tVec]_j}
\end{equation}


\begin{equation}\label{eq:Jacobian_part2}
    \matfont{H}_{\matfont{t}} = \matfont{R}_{cr'}
\end{equation}

%% file: Files/Equations/Jacobian_part2.tex
\begin{equation} \label{eq:Jtheta_1}
    \begin{split}
        \matfont{R}_{tmp, \alpha}  &= \matfont{R}_{cr'} \\
        \vecfont{b}_{tmp, \alpha} &= \matfont{rot}_y(\beta) \matfont{rot}_x(\gamma) \vecfont{t}_r = [b_x,b_y,b_z]^\textmd{T} \\
        \vecfont{c}_{tmp,\alpha} &= [b_x \textmd{sin}(\alpha)-b_y\textmd{cos}(\alpha), b_x \textmd{cos}(\alpha) -  b_y \textmd{sin}(\alpha), 0]^\textmd{T} \\
    \end{split}
\end{equation}

\begin{equation}\label{eq:Jtheta_2}
    \begin{split}
        \matfont{R}_{tmp,\beta}  &= \matfont{R}_{cr'} \matfont{rot}_z(\alpha) \\
        \vecfont{b}_{tmp,\beta} &= \matfont{rot}_x(\gamma) \vecfont{t}_r = [b_x,b_y,b_z]^\textmd{T} \\
        \vecfont{c}_{tmp,\beta}  &= [-b_x \textmd{sin}(\beta)+b_z\textmd{cos}(\beta),0, -b_x \textmd{cos}(\beta) -  b_z \textmd{sin}(\beta)]^\textmd{T} \\
    \end{split}
\end{equation}

\begin{equation}\label{eq:Jtheta_3}
    \begin{split}
        \matfont{R}_{tmp,\gamma}  &= \matfont{R}_{cr'} \matfont{rot}_z(\alpha) \matfont{rot}_y(\beta) \\
        \vecfont{b}_{tmp,\gamma} &= \vecfont{t}_r = [b_x,b_y,b_z]^\textmd{T} \\
        \vecfont{c}_{tmp,\gamma}  &= [0, -b_y \textmd{sin}(\gamma)-b_z\textmd{cos}(\gamma), b_y \textmd{cos}(\gamma) -  b_z \textmd{sin}(\gamma)]^\textmd{T} \\
    \end{split}
\end{equation}

\begin{equation}\label{eq:Jtheta_4}
    \matfont{H}_{\ThetaVec} = 
    \begin{bmatrix}
        \matfont{R}_{tmp,\alpha}\vecfont{c}_{tmp,\alpha},  
        \matfont{R}_{tmp,\beta}\vecfont{c}_{tmp,\beta},  
        \matfont{R}_{tmp,\gamma}\vecfont{c}_{tmp,\gamma}
    \end{bmatrix}
\end{equation}

%% file: Files/Equations/Jacobian_part3.tex
\begin{equation}\label{eq:Jacobian_obs_c}
    \matfont{H}_{cam} = 
    \begin{bmatrix}
    \frac{f_x}{z_c} & 0 & -f_x\frac{x_c}{{z_c}\cdot{z_c}} \\
    0 & \frac{y_c}{z_c} &  -f_y\frac{y_c}{{z_c}\cdot{z_c}}
    \end{bmatrix}
\end{equation}

%% file: Files/Equations/Jacobian_part4.tex
\begin{equation}\label{eq:Jacobian_final}
    \matfont{H} = \matfont{H}_{cam} \matfont{H}_{robot}
\end{equation}

%% file: Files/Tables/Nomenclature_JCBB.tex
\begin{table}[htbp]
\caption{Notation table for key points association}
\centering
\add{
\begin{tabular}{p{1.5cm} p{6.0cm}}
\toprule
\textbf{Symbol} & \textbf{Meaning} \\
\midrule
$\bar{\vecfont{x}}$ & ground truth state vector \\
$\check{\vecfont{x}}$ & predicted state vector (a priori) \\
$\bar{\vecfont{y}}$ & ground truth pixel projection \\
$\hat{\vecfont{y}}$ & observed pixel projection (a posteriori)\\
$\vecfont{e}$ & process noise \\
$\vecfont{v}$ & observation noise \\
$\mathbf{\Sigma_e}$ & process noise covariance \\
$\mathbf{\Sigma_v}$ & observation noise covariance \\
${\mathcal{P}_{pre}}$ & list of predicted pixel projections \\ 
${\mathcal{P}_{obs}}$ & list of predicted pixel projections \\
$\mathcal{H}_{ij}$ & hypothesis that ${\mathcal{P}_{obs,i}}$ is paired to ${\mathcal{P}_{pre,j}}$ \\
$\matfont{H}_{ij}$ & Jacobian that corresponds to $\mathcal{H}_{ij}$ \\
$\vecfont{h}_{ij}$ & innovation that corresponds to $\mathcal{H}_{ij}$ \\
$\matfont{C}_{ij}$ & covariance matrix for $\vecfont{h}_{ij}$ \\
$D_{ij}$ & squared Mahalanobis distance for $\mathcal{H}_{ij}$ \\
$\Hcand$ & joint hypotheses candidate \\
$\Hfinal$ & joint hypotheses output after JCBB check \\
$\matfont{C}_{\Hcand}$ & covariance matrix for $\Hcand$ \\
$D_{\Hcand}$ & squared Mahalanobis distance for $\Hcand$ \\
$l_{\Hcand}$ & negative logarithm of matching likelihood for $\Hcand$ \\
$\Hlist$ & list of joint hypotheses \\
\bottomrule
\end{tabular}
}
\label{tab:nomanclature_JCBB}
\end{table}

%% file: Files/Equations/JC.tex
\begin{align} 
    \label{eq:monte-model1}
    &\add{\vecfont{x}^{(t)} = p(\vecfont{x}^{(t-1)}, \vecfont{u}^{(t)}) + \vecfont{e}^{(t)}}, &\text{where } \add{\vecfont{e}^{(t)}} \sim \mathcal{N}(\vecfont{0}, \mathbf{\Sigma_e})  \\
    \label{eq:monte-model2}
    &\add{\vecfont{y}^{(t)} = g(\vecfont{x}^{(t)}) + \vecfont{v}^{(t)}},
    &\text{where } \add{\vecfont{v}^{(t)}} \sim \mathcal{N}(\vecfont{0}, \mathbf{\Sigma_v}) 
\end{align}

%% file: Files/Equations/ImplicitMeasure.tex
\begin{equation}\label{eq:implicity_measure}
    \begin{split}
    f(\add{\bar{\vecfont{x}}, \bar{\vecfont{y}})} &\approx \hat{\vecfont{y}} - g(\add{\check{\vecfont{x}}}) + \pd{f(\vecfont{x}, \vecfont{y})}{\vecfont{x}} (\add{\bar{\vecfont{x}}}-\add{\check{\vecfont{x}}})
    + \pd{f(\vecfont{x}, \vecfont{y})}{\vecfont{y}}(\add{\bar{\vecfont{y}}-\hat{\vecfont{y}}}) \\
    & = \vecfont{h}_{ij} + \matfont{H}_{ij}(\add{\bar{\vecfont{x}}-\check{\vecfont{x}}}) + (\add{\bar{\vecfont{y}}-\hat{\vecfont{y}}}) =  \vecfont{0}
    \end{split}
\end{equation}

%% file: Files/Equations/Cij_individual.tex
\begin{equation}\label{eq:C_mat_indi}
    \matfont{C}_{ij} = \matfont{H}_{ij} \mathbf{\Sigma_e} \matfont{H}_{ij}^\textmd{T} + \mathbf{\Sigma_v}
\end{equation}

\begin{equation}\label{eq:Mahalanobis}
    D_{ij}^2 = \vecfont{h}_{ij}^\textmd{T} \matfont{C}_{ij}^{-1} \vecfont{h}_{ij} < \chi^2_{d,\alpha}
\end{equation}

%% file: Files/Equations/Cij_joint.tex
\begin{equation}\label{eq:C_mat_joint}
    \begin{split}
        \matfont{C}_{\Hcand}=
        \begin{bmatrix}
            \matfont{H}_{i_1j_1}\\
            \vdots \\
            \matfont{H}_{i_kj_k}
        \end{bmatrix}
        &
        \begin{bmatrix}
            \mathbf{\Sigma_{e_1}} & & \\
            & \ddots & \\
            & & \mathbf{\Sigma_{e_k}} \\
        \end{bmatrix}
        \begin{bmatrix}
            \matfont{H}_{i_1j_1}\\
            \vdots \\
            \matfont{H}_{i_kj_k}
        \end{bmatrix}^\textmd{T} \\
        + &
        \begin{bmatrix}
            \mathbf{\Sigma_{v_1}} & & \\
            & \ddots & \\
             & & \mathbf{\Sigma_{v_k}} \\
        \end{bmatrix}
    \end{split}
\end{equation}

%% file: Files/Equations/Mahalanobis_joint.tex
\begin{equation}\label{eq:Mahalanobis_joint}
    D_{\Hcand}^2 = 
    \begin{bmatrix}
        \vecfont{h}_{i_1j_1}^\textmd{T} &\cdots& \vecfont{h}_{i_kj_k}^\textmd{T} \\
    \end{bmatrix}_{1\times2k}
    \matfont{C}_{\Hcand}^{-1}
    \begin{bmatrix}
        \vecfont{h}_{i_1j_1} \\
        \vdots \\
        \vecfont{h}_{i_kj_k}
    \end{bmatrix}_{2k\times1}
\end{equation}

%% file: Files/Equations/ML.tex
\begin{equation}\label{eq:ML}
    l_{\Hcand} = 2k\log(2\pi) + D^2_{\Hcand} + \log(\det(\matfont{C}_{\Hcand}))
\end{equation}

%% file: Files/Algorithms/JC_evaluation.tex
\begin{algorithm}
\caption{Joint Compatibility Evaluation}\label{alg:JC_algorithm}
\begin{algorithmic}[1]
\Require $\mathcal{P}_{pre}, \mathcal{P}_{obs}$, $\Hcand$
\Ensure $l_{\Hcand}$, $istrue$ \\
\textbf{Initialisation:} 
\\ $\mathbf{\Sigma_{e,all}}$, $\mathbf{\Sigma_{v,all}}$, $\matfont{H}_{all}$, $\vecfont{h}_{all}$ \Comment{Empty for initialisation}
\Procedure{\textbf{JC}}{$\Hcand$} \label{line:alg:JC_algorithm} 
\State $m \gets len(\Hcand)$
\If{$m=0$} \Comment{No observation}
    \State \textbf{return} $l_{\Hcand}= \textbf{inf}, istrue=\textbf{False}$
\Else
    \For{\text{all} \add{$\mathcal{H}_{ij}$} \textbf{in} $\Hcand$}
        \If{\add{$j=0$}} \Comment{\add{Outlier}}
            \State \textbf{continue}
        \Else
            \State \add{$\vecfont{h}_{ij}, \matfont{H}_{ij} \gets$} eq.\eqref{eq:Jacobian_part1} to \eqref{eq:Jacobian_final}
            \State $\mathbf{\Sigma_{e,all}}.add(\mathbf{\Sigma_{e_i}}), \mathbf{\Sigma_{v,all}}.add(\mathbf{\Sigma_{v_i}})$ 
            \State \add{$\matfont{H}_{all}.add(\matfont{H}_{ij}), \vecfont{h}_{all}.add(\vecfont{h}_{ij})$} 
        \EndIf
    \EndFor
    \State $\matfont{C}_{\Hcand}, D^2_{\Hcand}, l_{\Hcand} \gets$ eq.\eqref{eq:C_mat_joint}, \ref{eq:Mahalanobis_joint}, \ref{eq:ML}
    \State \textbf{return } $l_{\Hcand}, istrue = D^2_{\Hcand} < \chi^2_{d,\alpha}$
\EndIf

\EndProcedure
\end{algorithmic}
\end{algorithm}

%% file: Files/Algorithms/JCBB_evaluation.tex
\begin{algorithm}
\caption{Joint Compatibility Branch and Bound}\label{alg:JCBB_algorithm}
\begin{algorithmic}[1]
\Require $\mathcal{P}_{pre}, \mathcal{P}_{obs}$, \add{$\Hlist=\{\Hcand_1...\Hcand_m\}$}
\Ensure $\Hfinal$ \\
\textbf{Initialisation:} \\ $\Hfinal=\{\}$, $\Hcand=\{\}$, $n_{pair}=0$, $l=+ \infty$, $n=len(\mathcal{P}_{pre})$, $m=len(\mathcal{P}_{obs})$
\Procedure{\textbf{JCBB}}{$\Hcand$}\Comment{Starting point} \label{line:alg:JCBB_algorithm} 
\State \add{$i \gets len(\Hcand)$} \Comment{Count existing pairs}
\State $k_p \gets Pairs(\Hcand)$ \Comment{Count non-trivial pairs}
\If{$len(\Hcand)=m$}
    \State $l_{\Hcand}, istrue \gets \textbf{JC}(\Hcand, \mathcal{P}_{pre}, \mathcal{P}_{obs})$ \Comment{Alg.\ref{alg:JC_algorithm}}
    \If{($n_{pair}=0$ \textbf{or} $k_p>n_{pair}$) \textbf{and} $istrue$ }
        \State \hspace*{0.0em} \add{$\Hfinal \gets \Hcand, n_{pair} \gets k_p, l \gets l_{\Hcand}$} 
        \State \hspace*{0.0em} \textbf{return} \Comment{Prioritise larger $n_{pair}$}
    \ElsIf{\Statex \hfill $k_p=n_{pair}$ \textbf{and} $l_{\Hcand}<l$ \textbf{and} $istrue$}
        \State \hspace*{0.0em}  \add{$\Hfinal \gets \Hcand$, $l \gets l_{\Hcand}$} 
        \State \hspace*{0.0em} \textbf{return} \Comment{Prioritise smaller $l$}
    \Else  \textbf{ return}
    \EndIf
\Else
    \If{\add{$\Hcand_{i+1}=\{\}$}} \Comment{\add{Outlier}}
        \State $k_{max} \gets MaxRemPairs(\Hcand, \Hlist)$ \label{line:JCBB_pair_max_pair_1}
        \If{$k_{max}+k_p \geq n_{pair}$} \label{line:JCBB_pair_check_1}
            \State $\Hcand \gets \Hcand.add\{None\}$
            \State  \textbf{JCBB}($\Hcand$) \Comment{Case 1: to line \ref{line:alg:JCBB_algorithm}}
        \EndIf
    \EndIf
    \For{ \add{$\mathcal{H}_{(i+1)j} \textbf{ in }\Hcand_{i+1}$}}
        \State $k_{max} \gets MaxRemPairs(\Hcand, \add{\Hlist}, j)$ \label{line:JCBB_pair_max_pair_2}
        \If{$k_{max}+k_p \geq n_{pair}$} \label{line:JCBB_pair_check_2}
            \State \add{$\Hcand \gets \Hcand.add\{\mathcal{H}_{(i+1)j}\}$}
            \State $\textbf{JCBB}(\Hcand)$ \Comment{Case 2: to line \ref{line:alg:JCBB_algorithm}}
            
        \EndIf
    \EndFor
    \State $\Hcand \gets \Hcand.add\{None\}$
    \State \textbf{JCBB}$(\Hcand)$ \Comment{Case 3: to line \ref{line:alg:JCBB_algorithm}}
\EndIf

\EndProcedure
\State \textbf{return} $\Hfinal$\Comment{Final list of associations}
\end{algorithmic}
\end{algorithm}

%% file: Files/Equations/visibility_check.tex
\newcommand{\drfe}{d_{{rf},edge}}
\newcommand{\drbe}{d_{{rb},edge}}
\newcommand{\drre}{d_{{rr},edge}}
\newcommand{\drle}{d_{{rl},edge}}
\newcommand{\drfc}{d_{{rf},c}}
\newcommand{\drbc}{d_{{rb},c}}
\newcommand{\drrc}{d_{{rr},c}}
\newcommand{\drlc}{d_{{rl},c}}
\begin{equation} \label{eq:eta_fb}
    \eta_{fb} = \frac{\drfc + \drbc}{\drfe + \drbe} \geq \gamma  
 \end{equation}

\begin{equation} \label{eq:eta_lr}
    \eta_{lr} = \frac{\drlc + \drrc}{\drle + \drre} \geq \gamma   
\end{equation}

%% file: Files/Tables/Nomenclature_filter.tex
\begin{table}[htbp]
\caption{Notation table for the state update module}
\centering
\add{
\begin{tabular}{p{1.5cm} p{6.0cm}}
\toprule
\textbf{Symbol} & \textbf{Meaning} \\
\midrule
$\xpre$ & posterior state mean estimation at t-1 \\
$\sigxpre$ & posterior state covariance estimation at t-1 \\
$\xnowprior$ & a priori state mean estimation at t \\
$\xnowprior_{pt}$ & a priori particle state mean estimation at t \\
$\sigxnowprior$ & a priori state covariance estimation at t \\
$\vecfont{h}_{pt}$ & particle innovation \\
$\InnList$ & list of innovations \\
$\vecfont{r}_{ij}$ & residual for $\mathcal{H}_{ij}$ \\
$\alpha_f$ & forget factor \\
$w_{pt}$ & particle weight \\
$N_{eff}$ & number of effective particles \\
\bottomrule
\end{tabular}
}
\label{tab:nomanclature_filter}
\end{table}

%% file: Files/Algorithms/EKF.tex
\begin{algorithm}
\caption{Extended Kalman Filter }\label{alg:EKF_algorithm}
\begin{algorithmic}[1]
\Require $\xpre$, $\sigxpre$, $\Hfinal$
\Ensure \add{$\xnow$, $\sigxnow$}
\Procedure{\textbf{EKF}}{$\xpre$, $\sigxpre$, $\Hfinal$} \label{line:alg:EKF_algorithm} 
\State \add{$\xnowprior = \xpre$ \Comment{A priori estimation of \vecfont{x}}}
\State \add{$\sigxnowprior=\sigxpre + \mathbf{\Sigma_e}$ \Comment{A priori estimation of $\mathbf{\Sigma_x}$}}
\State $\InnList=\{\}$ \Comment{Innovation list}
\For{\add{$\mathcal{H}_{ij}$} \textbf{in} \add{$\Hfinal$}}
    \State \add{$\vecfont{h}_{ij} = [\hat{u}_j,\hat{v}_j]^\textmd{T}-[\check{u}_{i},\check{v}_{i}]^\textmd{T}$} \Comment{Innovation}
    \State \add{$\InnList.add\{\vecfont{h}_{ij}\}$}
    \State \add{$\matfont{C}_{ij}=\matfont{H}_{ij} \sigxnowprior \matfont{H}_{ij}^\textmd{T} + \mathbf{\Sigma_v}$} \Comment{eq.\eqref{eq:C_mat_indi}}
    \State \add{$\matfont{K}_{ij}= \sigxnowprior  \matfont{H}_{ij}^\textmd{T} \matfont{C}_{ij}^{-1}$}
    \State \add{$\xnowprior= \xnowprior + \matfont{K}_{ij} \vecfont{h}_{ij}$}
    \State \add{$\sigxnowprior = (\matfont{I} - \matfont{K}_{ij} \matfont{H}_{ij} )\sigxnowprior$}
\EndFor
\State \textbf{return} \add{$\xnow=\xnowprior, \sigxnow=\sigxnowprior, \InnList$}
\EndProcedure
\end{algorithmic}
\end{algorithm}

%% file: Files/Algorithms/AEKF.tex
\begin{algorithm}
\caption{Adaptive Extended Kalman Filter }\label{alg:AEKF_algorithm}
\begin{algorithmic}[1]
\Require $\Hfinal$, $\xpre$, $\sigxpre$, $\sigepre$, $\sigvpre$
\Ensure $\xnow$, $\sigxnow$, $\sigenow$, $\sigvnow$
\Procedure{\textbf{AEKF}}{$\Hfinal$} \label{line:alg:AEKF_algorithm} 
\State $\xnow$, $\sigxnow$, $\mathcal{P}_{inn}$ = \textbf{EKF}({$\xpre$, $\sigxpre$, $\Hfinal$})
\Comment{Alg.\ref{alg:EKF_algorithm}}
\State $m_k=Pairs(\Hfinal)$ \Comment{\add{Non-trivial pairs}}
\State \add{$\sigenow=\alpha_f \sigepre$, $\sigvnow=\alpha_f\sigvpre$}
\For{\add{$\mathcal{H}_{ij}$} \textbf{in} \add{$\Hfinal$}}
    \State \add{$\matfont{H}_{ij} \gets \text{Jacobian matrix evaluated at }\xnow$}
    \State \add{$\vecfont{h}_{ij} =\text{the $i^{th}$ element of } \mathcal{P}_{inn}$} \Comment{Innovation}
    \State \add{$\vecfont{r}_{ij} =  [u_{j},v_{j}]^\textmd{T}-g(\xnow)  $} \Comment{Residual}
    \State $\sigvnow = \sigvnow$ + \add{$\frac{1-\alpha_f}{m_k} (\vecfont{r}_{ij} \cdot \vecfont{r}_{ij}^\textmd{T} + \matfont{H}_{ij} \sigxpre \matfont{H}_{ij}^\textmd{T})$}
    \State \add{$\matfont{C}_{ij}=\matfont{H}_{ij} \sigxpre \matfont{H}_{ij}^\textmd{T} + \sigvpre$}
    \State \add{$\matfont{K}_{ij}= \sigxpre  \matfont{H}_{ij}^\textmd{T} \matfont{C}_{ij}^{-1}$}
    \State \add{$\sigenow = \sigenow + \frac{1-\alpha_f}{m_k} \matfont{K}_{ij} \cdot \vecfont{h}_{ij} \cdot \vecfont{h}_{ij}^\textmd{T} \cdot \matfont{K}_{ij}^\textmd{T}$}
\EndFor
\State \textbf{return} $\xnow, \sigxnow, \sigenow, \sigvnow$
\EndProcedure
\end{algorithmic}
\end{algorithm}

%% file: Files/Algorithms/PF.tex
\begin{algorithm}
\caption{Particle Filter }\label{alg:PF_algorithm}
\begin{algorithmic}[1]
\Require $\xpre$, $\sigxpre$, $\Hfinal$, $N_p$, $N_{eff}$, $\mathcal{P}_{obs}$
\Ensure $\xnow$, $\sigxnow$
\Procedure{\textbf{PF}}{$\xpre)$, $\sigxpre$, $\Hfinal$} \label{line:alg:PF_algorithm} 
\For{all particles}
    \State \add{$\xnowprior_{pt}$} = $Gaussian( \xpre, \sigxpre )$
    \State \add{$|\vecfont{h}_{pt}||_2 \gets $} update from \add{$\xnowprior_{pt}$}
    \State \add{$w_{pt} = \frac{1}{|| \vecfont{h}_{pt}||_2} $} \Comment{weight update}
    \EndFor
\State particle weight list $\{w_{pt}\}$ normalization
\State $\xnow \gets $  \add{$\sum \xnowprior_{pt} \cdot  w_{pt}$}, across all $N_p$ particles

\State stratified resampling \textbf{if} $\frac{1}{\sum w^2_{pt}}<N_{eff}$

\State \textbf{return} $\xnow$, $\sigxnow = \sigxpre$
\EndProcedure
\end{algorithmic}
\end{algorithm}

%% file: Files/Experiments.tex
\section{Experiments and results} \label{sec:experiments}
\subsection{Dataset and hardware setup}
The proposed algorithm framework was evaluated on different publicly available datasets, including multiple video recordings from both the SuPer dataset \footnote{\url{https://sites.google.com/ucsd.edu/super-framework/home}} and the SurgPose dataset \cite{wu2025surgpose}. These datasets contain demonstrations of various da Vinci instruments performing either \textit{in-vitro} tasks on tissue phantoms or \textit{ex-vivo} tasks on animal organs, with raw kinematics data recorded directly from the encoders. All codes for algorithm implementation and evaluation were written in Python 3.10 and run on HP Z2 Tower G9 workstation without GPU acceleration.


\subsection{Evaluation of JCBB data association}
\begin{figure*}[h]
    \centering
    \includegraphics[width=\textwidth]{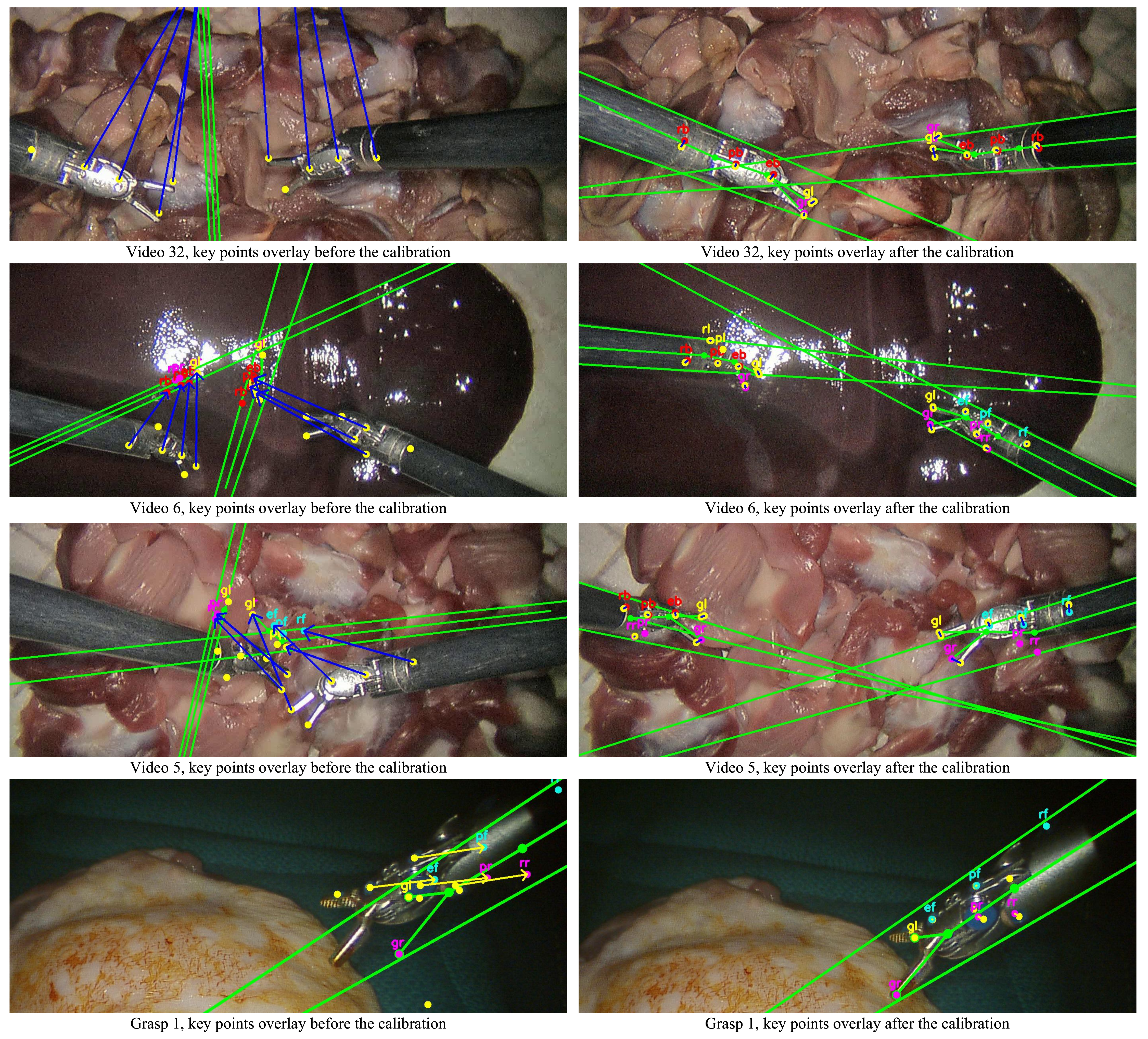}
    \caption{Illustrations of calibration performance. Overlays of key points, tool edges, and the tool skeleton are used to illustrate calibration performance. They show that, even in the presence of large initial calibration errors, reflected by disorientated and out-of-view tool overlays, the JCBB algorithm is still capable of establishing feature associations, resulting in effective calibration. It is noticed that, despite using different calibration approaches, the skeleton of the left instrument after calibration in video 5 does not align with its visual observation, indicating that the estimated RCM position is incorrect.}
    \label{fig: calibration illustration}
\end{figure*}

Video 32 from the SurgPose dataset was first used to validate the effectiveness of the JCBB feature association and visibility algorithm blocks. The video contains a total of 1,001 frames, in which two Long Needle Drivers (LND), attached to PSM1 and PSM3 respectively, operate on chicken gizzard. Illustrations of calibration performance are shown in Fig.\ref{fig: calibration illustration}. In all video recordings within the SurgPose dataset, key points are marked on instruments using ultraviolet reactive paint and are labeled from 1 to 14, where labels 1-7 correspond to the PSM1 instrument, and labels 8-14 correspond to the PSM3 instrument. However, this labeling convention does not distinguish the side of the instrument to which each key point belongs. For example, key point 1 may refer to either ``rb" or ``rf" under the convention defined in this paper. Additionally, in some frames, key points 6 and 7 correspond to ``pl/pr" and ``rl/rr", respectively, while in other frames these correspondences are reversed. Therefore, prior to validation, the key points were manually relabeled to match the convention adopted in this paper. Video 32 was selected because it features minimal instrument rotation, with one dominant instrument visible to the camera for each tool, which simplifies the relabeling process. The initial hand-eye calibration matrix was calculated via the RANSAC PnP algorithm \cite{RANSAC_PNP}, using manually labeled 2D-3D correspondences from the first one hundred frames. 

\begin{figure*}[h]
    \centering
    \includegraphics[width=\textwidth]{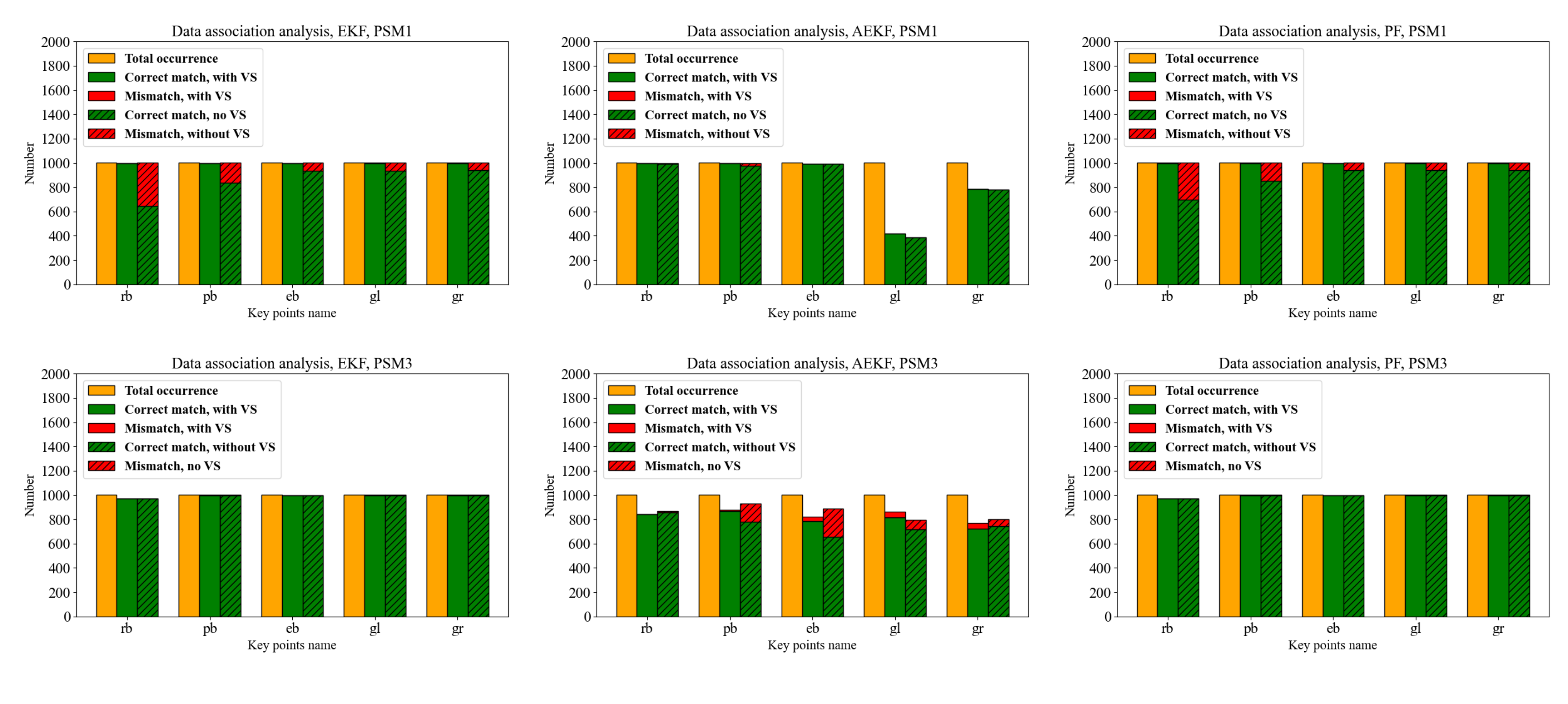}
    \caption{SurgPose 32, feature association analysis. The three images in the top row and the three images in the bottom row show the key points association results of three estimators for PSM1 and PSM3, respectively.}
    \label{fig:Association init 100}
\end{figure*}

Key points feature association results using the JCBB algorithm with different estimators are presented in Fig.\ref{fig:Association init 100}. The results demonstrate the high key points association accuracy achieved by the JCBB algorithm accompanied by visibility check. As shown in Fig.\ref{fig:Association init 100}, all labeled key points belonging to the PSM1 and PSM3 instruments appeared in all 1,001 frames. During data analysis, three scenarios were considered: (i) correct match, where the observed key point was matched to the correct label, (ii) mismatch, where the observed key point was matched to an incorrect label, and (iii) non-match, where the observed key point was omitted from the data association, such case not shown in the bars. It is observed that, with the inclusion of the visibility check block, the accuracy of feature association increased for all estimator approaches. This is primarily because after the visibility check, at least half of the candidate key points were removed, which effectively reduced the number of outliers in the candidate group. It is also noted that, compared to the EKF and PF approaches, the AEKF approach produced the largest number of non-match cases. This can be explained by the nature of the AEKF, which particularly adjusts the state and measurement covariances on-the-fly. These varying covariance matrices can affect the JCBB feature association results by influencing the computation of both the individual and joint Mahalanobis distances between candidate key point pairs. Consequently, key points associations established under static covariances may no longer satisfy the required thresholds once the variances are changed, leading to an increased number of non-match cases. However, the rise of non-matches does not necessarily compromise the overall calibration accuracy, as shown in Fig.\ref{fig:Errors 3D, visibility, surg32}; on the contrary, it can significantly accelerate the entire algorithm pipeline. Fig.\ref{fig:Time comparison, visibility, surg32} shows that, with the visibility check, the overall algorithm implementation time was significantly reduced for all estimators. For the EKF and AEKF approaches, the filter implementation time was negligible compared to the feature association time. In contrast, for the PF approach, the filter implementation time was substantial and was determined by the number of selected particles.

\begin{figure*}[h]
    \centering
    \includegraphics[width=\textwidth]{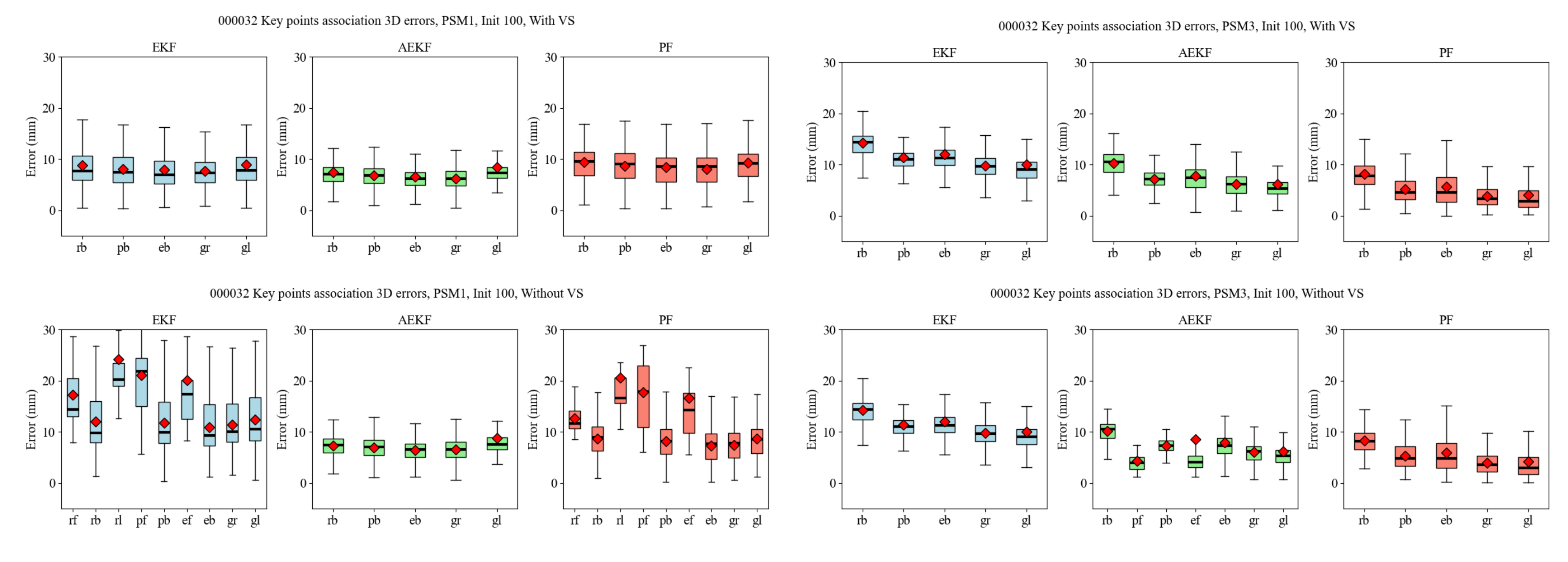}
    \caption{SurgPose video 32, 3D errors comparison. The two images in the top row show 3D errors for associated key points with the visibility check, while the two images in the bottom row show 3D errors without VS. It is observed that, without the visibility check, an increased number of incorrect key points associations were found, resulting in larger 3D errors.}
    \label{fig:Errors 3D, visibility, surg32}
\end{figure*}

\begin{figure*}[h]
    \centering
    \includegraphics[width=\textwidth]{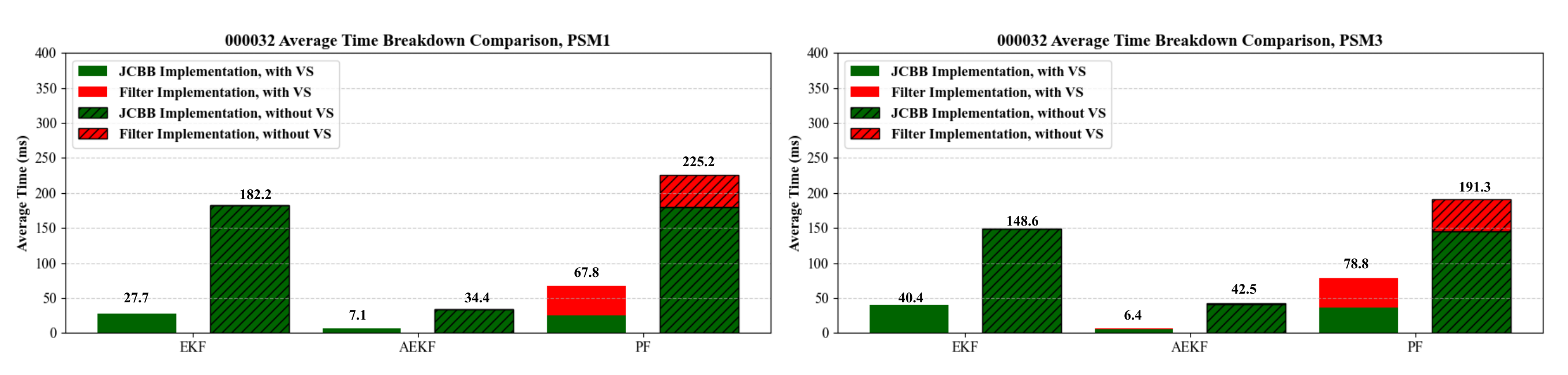}
    \caption{SurgPose video 32, time comparison. Compared to the EKF and AEKF approaches, the PF approach required more time for filter implementation. With the visibility check, the overall implementation time for all filter approaches was significantly reduced.}
    \label{fig:Time comparison, visibility, surg32}
\end{figure*}

\subsection{Evaluation of 3D tracking performance}
Based on the key points association results, pixel discrepancies are fed into estimators, which in turn produce estimates of the calibrated hand-eye transformation matrix. Subsequently, the 3D positions of these key points can be estimated through forward kinematics and the calibrated hand-eye transformation matrix. The SurgPose dataset provides pixel coordinates of labeled key points on stereo images, as well as intrinsic camera parameters. Therefore, the ground-truth 3D position of all labeled key points can be obtained via triangulation implemented in OpenCV \cite{opencv_library}. The reconstructed 3D positions of associated key points using different estimators are compared and analyzed. The performances of EKF, AEKF, and PF are also compared with that of the RANSAC PnP algorithm, which incrementally takes input as all 2D-3D key points correspondences from the past history. Video 5 was chosen for evaluation, featuring two LNDs in operation. To examine the impact of initial calibration errors, a set of initializations with varying levels of accuracy were prepared. To achieve this, the number of frames used for the initial calibration was set to 10, 50, 100 and 200, respectively. The results for key points 3D position reconstruction are presented in Fig.\ref{fig: Surg5, 3D tracking performance}. It shows that, with an increase of the initial frames, the 3D reconstruction errors for all key points on the PSM1 instrument decreased for all estimators, despite still being less accurate than the PnP method. This implies that these estimators are sensitive to initial calibration errors. The results also suggest that, for the AEKF approach, key point mismatches occurred when the number of initial frames was not sufficient, as reflected by the two additional key point categories compared to the other approaches. In contrast, for the PSM3 instrument, the 3D reconstruction accuracy did not improve significantly with more calibration frames, implying that the initial calibration result derived from the first 10 frames was sufficiently accurate. It is also noteworthy that the reconstruction accuracy of the estimator approaches exceeded that of the PnP approach, suggesting the advantage of estimator-based approaches when initial calibration errors are small. It is suspected that, the initial calibration accuracy for the PSM3 instrument was higher than that for the PSM1 instrument, despite using the same number of frames, because more visible key points were located on different sides of the PSM3 instrument, which made the initial calibration result more robust across the trial.

\begin{figure*}[h]
    \centering
    \includegraphics[width=\textwidth]{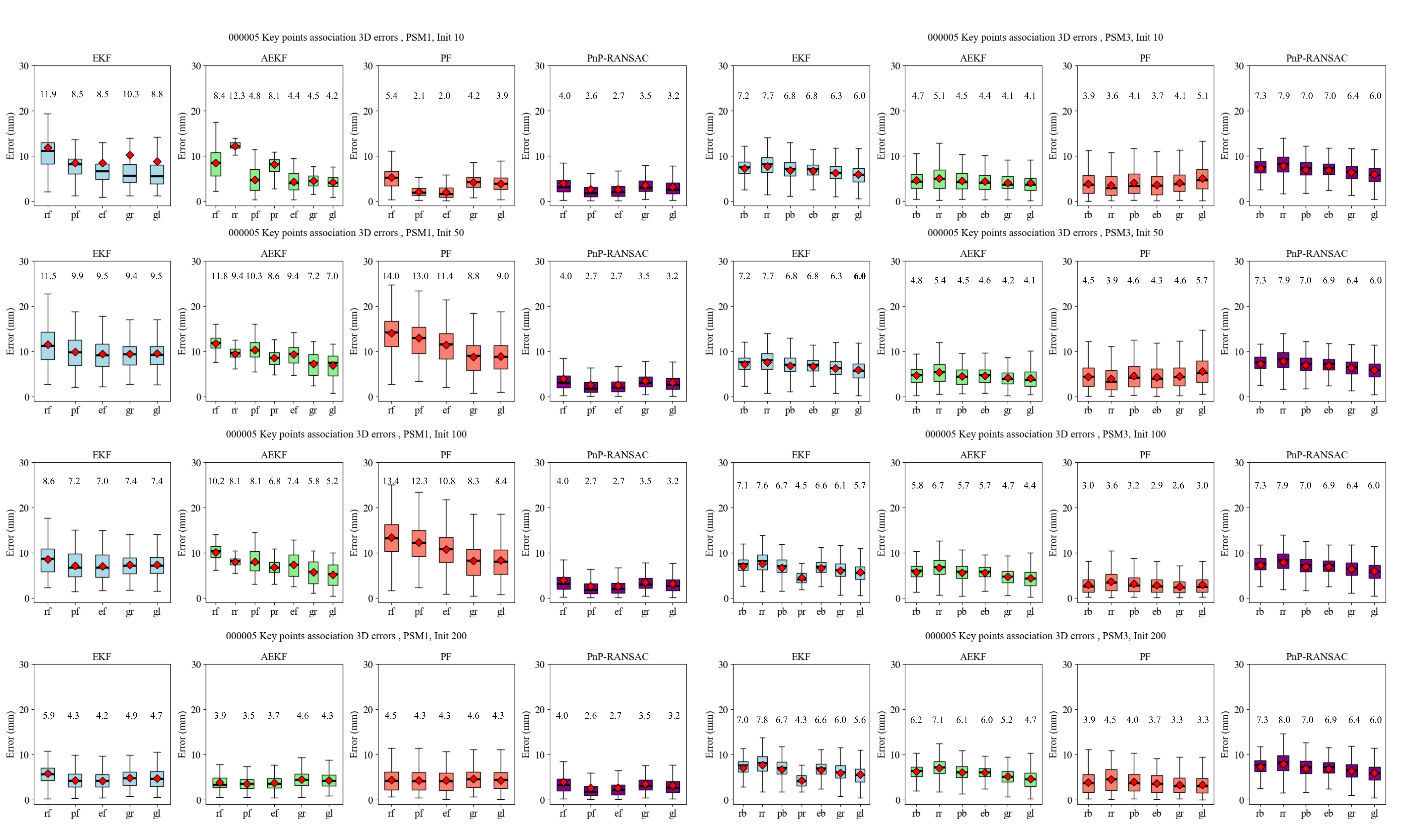}
    \caption{SurgPose video 5, Initial calibration frame analysis. Mean errors are labeled on the images. ANOVA tests were conducted on the key point errors produced by different approaches, and all p-values were all smaller than $10^{-5}$.}
    \label{fig: Surg5, 3D tracking performance}
\end{figure*}

\subsection{Evaluation of robustness against disturbances}
To evaluate the performance of different estimator-based approaches under sudden disturbances to the hand-eye transformation matrix, as occurs when the camera is abruptly repositioned, random disturbances of varying magnitudes were added to the hand-eye calibration matrix every 25 frames. Disturbances are categorized as ``low", ``medium", and ``high". The ``low" level applies random rotation noises of $\pm1$ degrees to all three rotation components, and random translation noises of $\pm1$ cm to all three translation components. Similarly, the ``medium" level applies random rotation noises of $\pm3$ degrees and translation noises of $\pm3$ cm, and the ``high" level applies random rotation noises of $\pm5$ degrees and random translation noises of $\pm5$ cm. Video 6 from the SurgPose dataset was selected, featuring two LNDs operating on a pork liver, and the number of initial calibration frames was set to 100. Fig.\ref{fig: Surg6, errors 2D PSM} shows the 2D pixel association errors when no disturbances were added. Fig.\ref{fig: Surg6, disturbance analysis} shows the performance of different estimators under random high-level disturbances. It shows that, compared to when no disturbances were added, larger 3D errors were found for all three estimator approaches, whereas the PnP method produced more consistent results. This difference can be attributed to the fact that estimators require sequential frames to gradually reduce calibration errors, whereas the PnP method relies only on the current frame and historical correspondences. Similar results were obtained for cases when ``low" and ``medium" levels of disturbances were added.

\begin{figure*}[h]
    \centering
    \includegraphics[width=\textwidth]{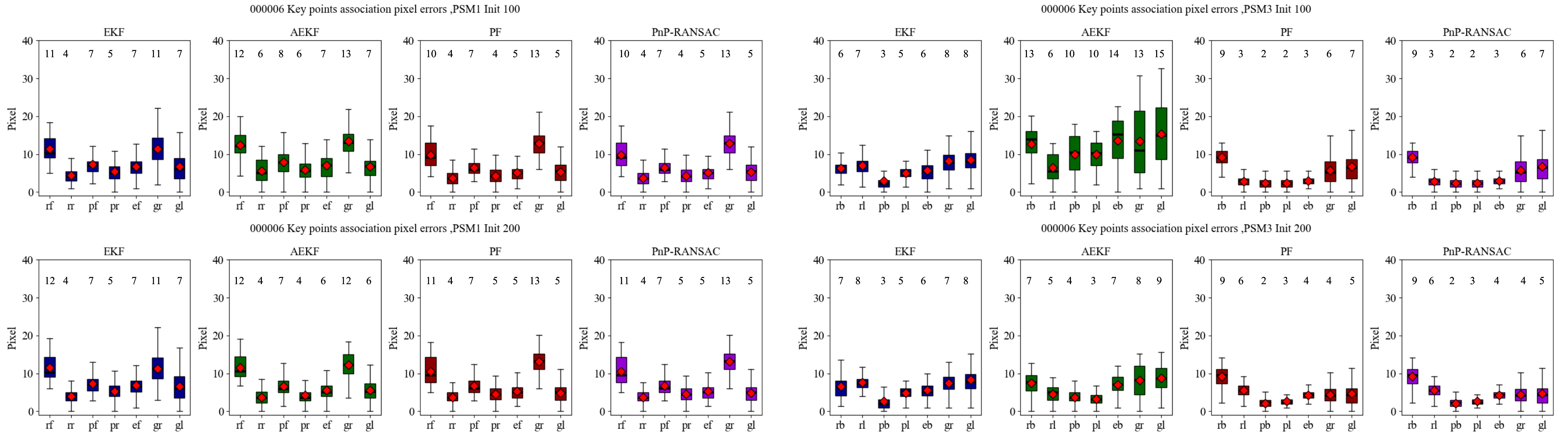}
    \caption{SurgPose video 6, 2D pixel errors. The 2D key point pixel association errors for the PSM1 and PSM3 instruments, with different initial calibration frames, are presented. It is noticed that there is a significant decrease in PSM3 pixel errors for AEKF approach as the number of initial calibration frames rises to 200.}
    \label{fig: Surg6, errors 2D PSM}
\end{figure*}

\begin{figure*}[h]
    \centering
    \includegraphics[width=\textwidth]{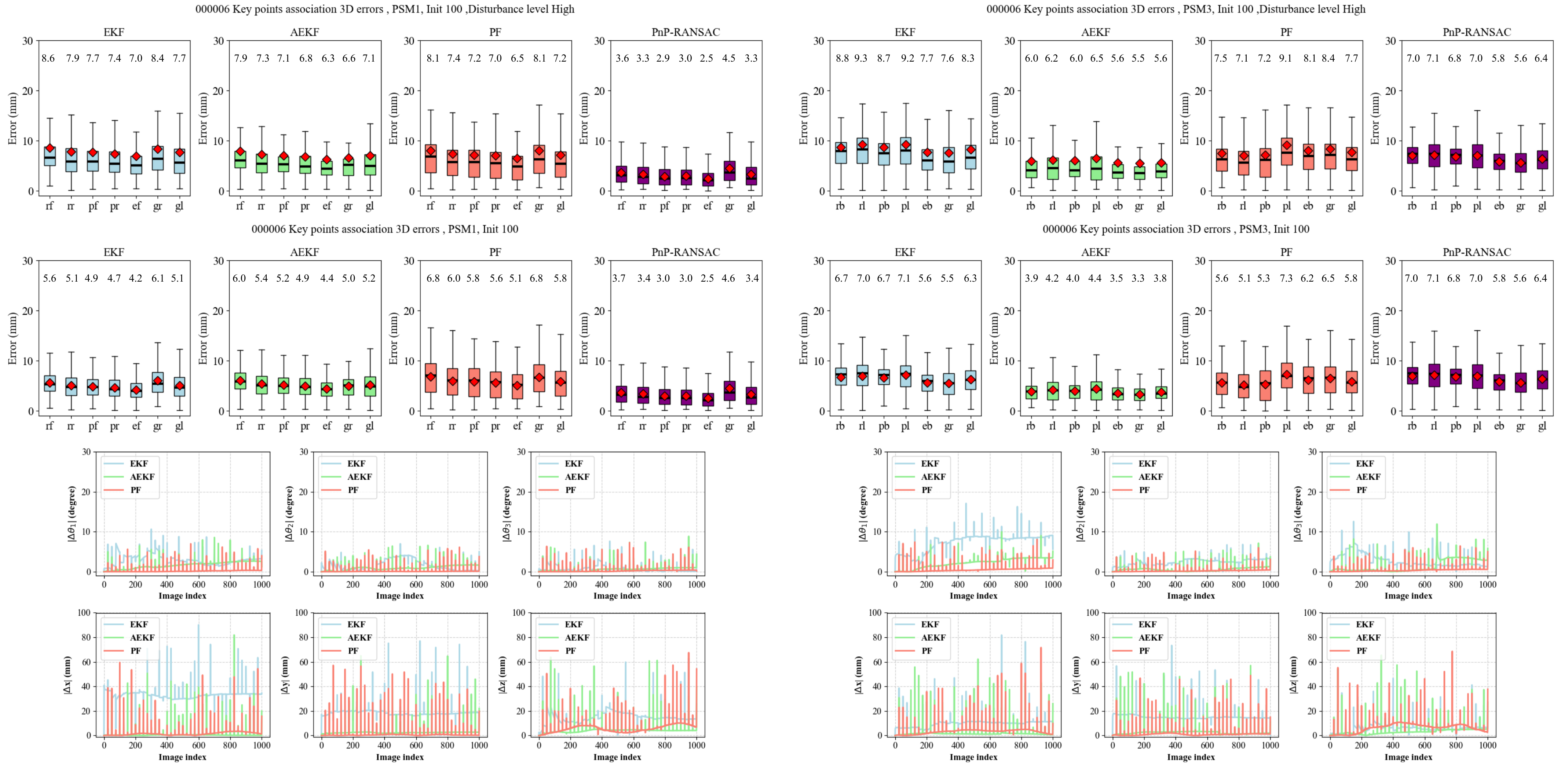}
    \caption{SurgPose video 6, disturbance analysis, high level. The 3D reconstruction errors for all key points on the PSM1 and PSM3 instruments, with and without applied disturbances, are presented. Mean errors are labeled in the images. The two images in the bottom row show the trends of the six variables that constitute the on-the-fly calibrated hand-eye transformation matrix for the PSM1 and PSM3 instruments, respectively. Spikes appear every 25 frames, matching the frequency at which high-level disturbances were introduced.}
    \label{fig: Surg6, disturbance analysis}
\end{figure*}

\subsection{Evaluation of robustness against measurement errors}
Performances of different estimators were also assessed on the SuPer dataset. Unlike the SurgPose dataset, where ground truth pixel coordinates of key points are directly provided, in the SuPer dataset, key points on the instrument are labeled using blue markers, and their pixel coordinates are extracted via HSV color masking, which introduces more measurement errors. As there lacks ground truth results for the 3D key points positions, the performance of three estimators are compared with that of the PnP method. Comparison of the reconstructed hand-eye calibration matrix on different video datasets are shown in Table.\ref{tab:Tcr_Super}. 

\input{Files/Tables/SuperDataset_Tcr_comparison}

\subsection{\add{End-to-end pipeline evaluation}}
\begin{figure*}[h]
    \centering
    \includegraphics[width=\textwidth]{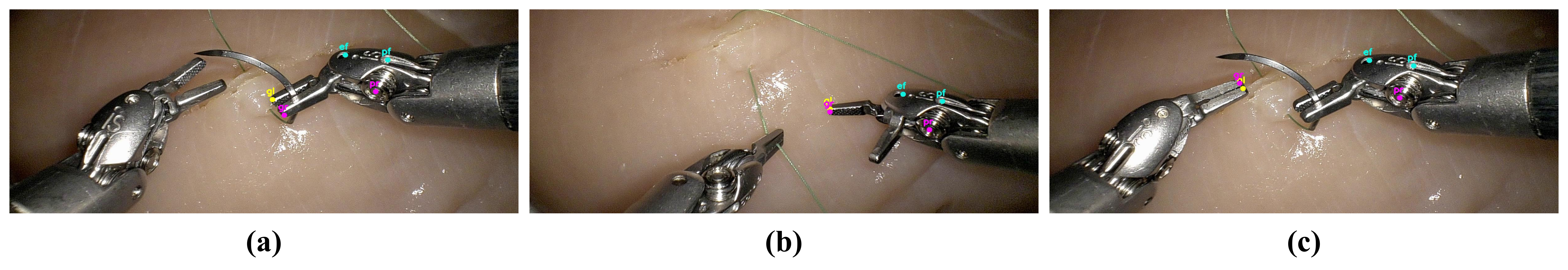}
    \caption{\add{Single-instrument key point detection using a fine-tuned DeepLabCut model. (a) shows correct detection results; (b) shows false positive in which the key points ``gl" and ``gr" are confused; (c) shows false positive in which the key points ``gl" and ``gr" are incorrectly detected on the other instrument.}}
    \label{fig: false detections}
\end{figure*}

\begin{figure*}[h]
    \centering
    \includegraphics[width=\textwidth]{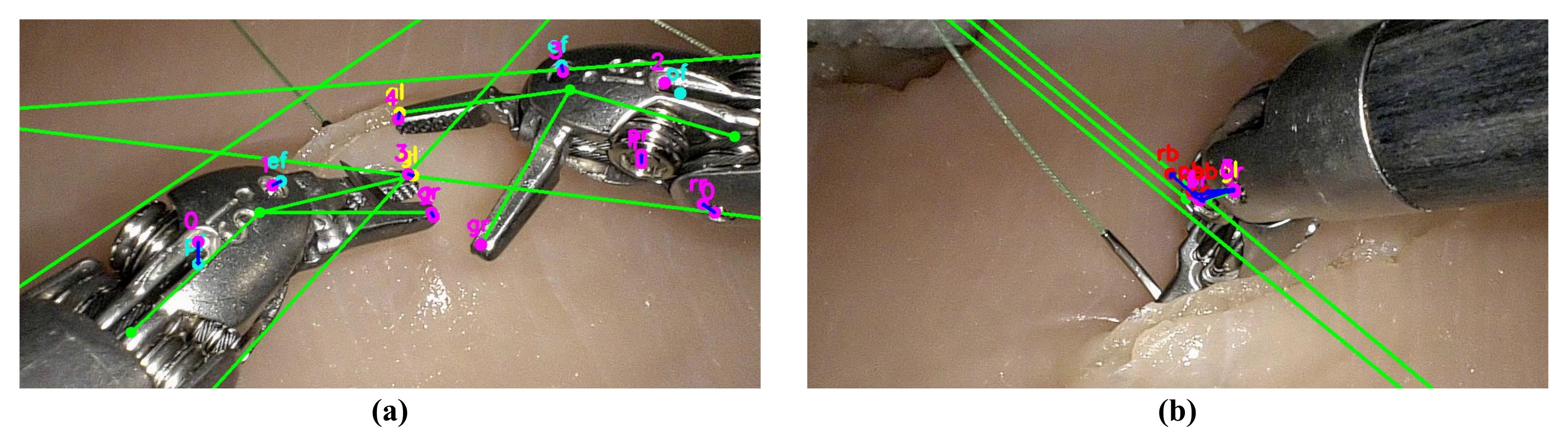}
    \caption{\add{(a) shows the visual alignment result obtained using the Particle Filter approach; (b) illustrates a common scenario in which the PnP approach fails due to limited feature detections and false associations. Numbers from 0 to 4 indicate key points detected using a fine-tuned DeepLabCut model.}}
    \label{fig: failure pnp}
\end{figure*}

\add{In this subsection, we further evaluate the performance of the proposed calibration framework by incorporating a learning-based key point detection module to allow for scenarios where temporally stable detections cannot be guaranteed, which is less common for marker-based approaches. The DeepLabCut model \cite{DLC3_rebuttal} was adopted and fine-tuned on \textit{ex vivo} frames curated by Johns Hopkins University as part of the Open-H-Embodiment dataset \cite{nelson2026openH_nvidia}. Altogether, 134 frames featuring two needle instruments performing suturing tasks were automatically selected and then manually labeled in the GUI toolbox as a multi-animal project. During annotation, only visible key points were labeled, and no distinction was made between key points located on symmetrical sides of the instrument. Performance was evaluated on 204 unseen frames. Although key point detections contain class labels, they are not consistently reliable, as demonstrated in Fig.\ref{fig: false detections}. The first five frames with correctly predicted labels were used to estimate the initial hand-eye transformation. Unlike in previous experiments, all filter-based approaches demonstrated effective tool skeleton tracking, as evidenced by visual alignment results. In contrast, the PnP method exhibited instability and failures, particularly when few or no key points were detected, as illustrated in Fig.\ref{fig: failure pnp}. This can be explained by the fact that the PnP method contains no memory of previous estimations, resulting in more drastic changes across frames. Consequently, it is more susceptible to spurious detections and false associations. When a false estimation is generated, it adversely affects subsequent feature associations, leading to increasingly erroneous estimations that eventually accumulate and result in complete failure.}

%% file: Files/Tables/SuperDataset_Tcr_comparison.tex
\begin{table*}[htbp]
\caption{Comparison of calibrated $\matfont{T}_{cr}$ results on different video datasets. $\matfont{T}_{cr}$ values obtained from estimator-based approaches are compared with those produced by the PnP-RANSAC method.}
\centering
\begin{tabular}{p{3.8cm} | p{1.8cm} | p{1.8cm} | p{1.8cm} | p{1.8cm} | p{1.8cm} | p{1.8cm}}
\toprule
\textbf{Dataset} & $|\Delta \vecfont{t}|$ \textbf{(mm)} \newline EKF & $|\Delta \vecfont{t}|$ \textbf{(mm)} \newline AEKF & $|\Delta \vecfont{t}|$ \textbf{(mm)} \newline PF  & $|\Delta \vecfont{r}|$ \textbf{(rads)} \newline EKF & $|\Delta \vecfont{r}|$ \textbf{(rads)} \newline AEKF & $|\Delta \vecfont{r}|$ \textbf{(rads)} \newline PF \\
\midrule
SuPer Dataset Grasp 1 & $2.74\pm1.13$ & $2.65\pm1.30$ & $14.71\pm4.14$ & $0.09\pm0.03$ & $0.14\pm0.04$ & $0.18\pm0.02$ \\
SuPer Dataset Grasp 2 & $4.51\pm4.12$ & $8.52\pm2.31$ & $16.55\pm2.46$ & $0.10\pm0.04$ & $0.14\pm0.02$ & $0.18\pm0.02$  \\
SuPer Dataset Grasp 3 & $11.34\pm3.63$ & $5.55\pm1.85$ & $16.24\pm3.80$ & $0.14\pm0.03$ & $0.21\pm0.02$ & $0.19\pm0.03$  \\
SuPer Dataset Grasp 4 & $6.67\pm6.04$ & $7.06\pm5.73$ & $21.43\pm6.63$ & $0.23\pm0.32$ & $0.21\pm0.31$ & $0.27\pm0.31$  \\
SuPer Dataset Grasp 5 & $5.91\pm9.12$ & $7.63\pm9.06$ & $20.04\pm8.93$ & $0.23\pm0.28$ & $0.21\pm0.28$ & $0.27\pm0.28$  \\
Surg 5, PSM1, init 200 & $13.90\pm14.40$ & $14.90\pm14.02$ & $14.19\pm14.52$ & $0.14\pm0.13$ & $0.20\pm0.13$ & $0.23\pm0.13$ \\
Surg 5, PSM3, init 200 & $2.98\pm2.47$ & $3.04\pm2.97$ & $6.82\pm3.38$ & $0.02\pm0.02$ & $0.02\pm0.02$ & $0.02\pm0.02$ \\
Surg 6, PSM1, init 100 & $3.12\pm5.28$ & $4.11\pm4.69$ & $6.15\pm5.17$ & $0.03\pm0.04$ & $0.04\pm0.04$ & $0.07\pm0.04$ \\
Surg 6, PSM3, init 100 & $2.84\pm3.37$ & $8.19\pm4.00$ & $16.22\pm6.86$ & $0.03\pm0.03$ & $0.07\pm0.03$ & $0.09\pm0.03$ \\
\bottomrule
\end{tabular}
\label{tab:Tcr_Super}
\end{table*}

%% file: Files/Discussions.tex
\section{Discussions} \label{sec:discussions}
Through extensive experiments on different video datasets, the effectiveness of the JCBB key point association algorithm, together with the visibility block, were validated. \add{For both} estimator-based calibration approach and direct PnP approach, reliable key point association is crucial towards obtaining accurate on-the-fly hand-eye calibration. The incorporation of the visibility check block not only reduces the number of outliers in the candidate list, thereby increasing feature association accuracy, but also accelerates the overall pipeline. \add{Although the experiments were conducted offline on a CPU, the framework is readily applicable to GPU-based implementations, where parallel processing and the decoupling of the AI inference thread could further support real-time video processing.} The practicability of key point association-based calibration approaches is also supported by the versatility of key point detection methods. For example, by applying infrared fluorescent dye onto instrument key points, their pixel coordinates can be directly detected in the Firefly mode using the dVRK Xi \cite{STIR_Schmidt}, without the requirement for pre-training. \add{Conversely, variants of the DeepLabCut framework reported in the literature support inference speeds of up to 100 Hz \cite{DLC_100HZ_latency}.} Additionally, this procedure is generalizable to different surgical instruments on different surgical platforms as long as their CAD models are available.

\begin{figure}[h]
    \centering
    \includegraphics[width=0.4\textwidth]{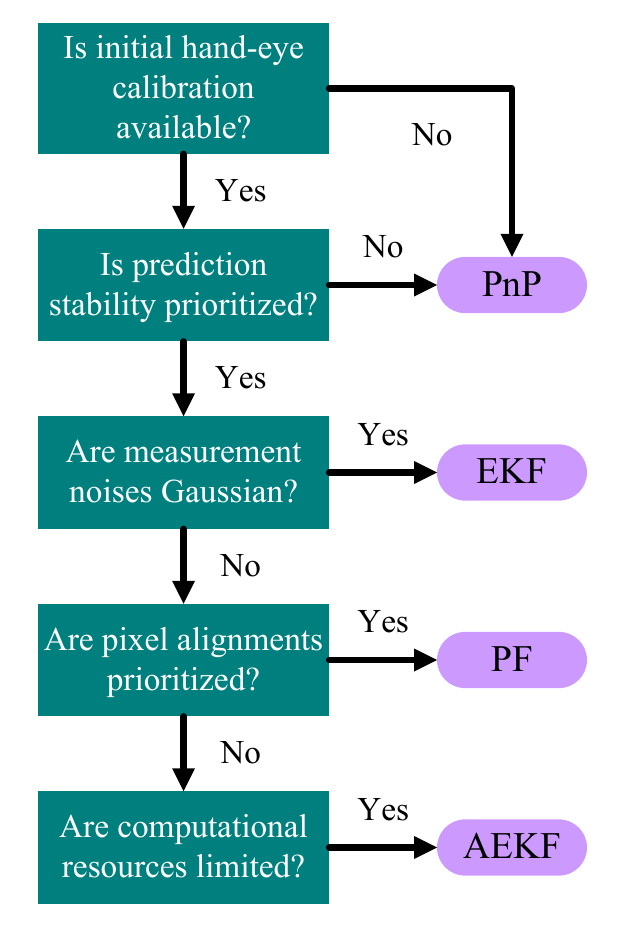}
    \caption{\add{Decision tree for calibration method selection.}}
    \label{fig: Filter Decision}
\end{figure}

The performance of different key point association-based approaches was evaluated under challenges including initial hand-eye calibration errors, random intra-operative disturbances, and key point measurement errors. The results show that estimator-based approaches are more susceptible to the accuracy of initial calibration, \add{whereas the PnP approach is sensitive to the accuracy of intra-operative feature detection and association.}
Table.\ref{tab:Tcr_Super} compares the estimated $\matfont{T}_{cr}$ between estimator-based approaches and the PnP-RANSAC method and shows significant discrepancies in the translational component. It can be explained from the observation that the PnP method tends to generate less consistent calibration results, as reflected by rapid changes in tool overlay, as shown in the videos included in the supporting material. \add{It is also noted that the PnP method retains no memory of its previous estimations, thereby enabling itself to more prompt calibrations on $\matfont{T}_{cr}$. In contrast, estimator-based approaches are suitable for consistent calibrations on encoder drifts over time. Therefore, when accurate feature detections and associations are available, the PnP method is more efficient, whereas estimator-based approaches are more robust in providing consistent results under conditions of temporal instability and limited feature detections.} \add{Among the three filter-based approaches, EKF is limited to scenarios where observation errors are Gaussian; PF is suitable for extremely noisy environments where sufficient features can be observed, and AEKF is more widely applicable to situations with limited computational resources.} \add{A decision tree is presented in Fig.\ref{fig: Filter Decision} to assist method selection.}

Compared with other estimator-based approaches that use key point associations, Ye \etal \cite{MengLong_2016_MICCAI} reported an average pose reconstruction error of $2.83\pm2.19$ (mm) in translation, and $0.13\pm0.10$ (rads) in rotation, by using their EKF framework. Although no ground truth values for the instrument pose were provided in the original datasets, a common practice to find the ground truth pose is to run the PnP algorithm after manually labeling 2D key points on the image. Therefore, Table.\ref{tab:Tcr_Super} provides a rough metric for comparison. The large discrepancy in $\Delta \vecfont{t}$ was caused by the fact that the estimators solely focused on reducing pixel association errors but failed to account for the RCM constraint. This scenario is reflected by a mirrored tool skeleton overlay on the image, as shown in videos included in the supporting material. On the other hand, when the accuracy of 3D depth estimation takes priority, estimators that rely on key points can outperform those that rely on differential rendering and silhouette matching \cite{Hao_silhouette_rendering, liang2025differentiable, yang2025real} in accuracy and speed.   

Compared with other recent pose estimation methods that do not rely on estimators or the PnP method, Fan \etal \cite{fan2024reinforcement} reported an average error of $2.81$ mm for 3D instrument joint position estimation on their semi-synthetic dataset, by using a reinforcement-learning based approach. This level of accuracy could also be achieved using estimator-based approaches on certain datasets, as shown in Fig.\ref{fig: Surg5, 3D tracking performance}. Although the JCBB feature association and visibility check algorithms were originally proposed for estimators-based approaches, they can also be valuable for other tool pose estimation algorithms, because they remove ambiguity in key points detected on the two symmetrical sides of the instrument, which is important when large articulations occur. Additionally, they can also be helpful in establishing inter-frame key point correspondences in an efficient way, which is an essential component in \cite{yang2025instrument}. 

Since there also exist studies that address the pose tracking problem via direct calibration of joint angles \cite{Berkeley_automation_Hwang, li2024real}, it would be useful to conduct comparisons in the joint space. As shown in \cite{Berkeley_automation_Hwang}, given the end-effector pose and the 3D position of joint 4 in the camera frame, joint angles 1-6 can be computed analytically. However, due to a lack of ground-truth instrument poses in the datasets, joint-space comparisons are difficult to evaluate in this study. To enable more extensive evaluations, it is hoped that future datasets will provide ground-truth results for both the instrument pose and the robot joint angles. 

%% file: Files/Conclusions.tex
\section{CONCLUSIONS} \label{sec:conclusions}
In this work, we proposed a generalizable on-the-fly hand-eye calibration framework with the aim of improving pose estimation/tool localization accuracy for the da Vinci surgical instruments, leveraging both raw kinematics data and key point features from monocular images. The framework consists of a JCBB feature association algorithm block and a calibration algorithm block. Without relying on any pre-training, the JCBB algorithm proved capable of establishing associations between 2D key points and their corresponding 3D positions with high accuracy, even in the presence of outliers. The performance of the JCBB algorithm was shown to be further enhanced with the fusion of the visibility check block, which effectively removed invisible key points from the candidate list. Given the established 2D-3D correspondences, several options were made available for the calibration algorithm block, including the EKF, AEKF, PF, and the PnP method. We provided a pool of calibration algorithms out of the concern that a single approach may fail to accommodate the different noise distributions that arise in various different surgical scenarios, such as large initial errors, sudden disturbances, and large measurement errors.

Through extensive experiments on different publicly available video datasets, the effectiveness of the proposed framework was validated and compared with other approaches. One issue identified was that the accuracy of the translational component of the estimated hand-eye calibration matrix remained inconsistent. This was suspected to be caused by the fact that estimator-based approaches prioritized reducing 2D association errors without accounting for physical constraints. \add{A potential solution is to incorporate the RCM constraint into the framework, using the reconstructed 3D RCM position as a pseudo-observation. To account for the disparity in error scales between pixel-based observations and the 3D pseudo-observation, a dynamic weighting factor can be introduced.} Additionally, we intend to test the proposed framework in more clinically realistic scenarios and expand the analysis to the joint space. Although the entire framework was programmed in Python for fast prototyping, to allow for better time performance, it will be rewritten in C++ and released as open source.

%% file: Files/Appendix.tex
\section{APPENDIX}
\subsection{Parameter selection}
Parameters used in the experiments are listed in Table.\ref{tab:parameters_JCBB} and \ref{tab:parameters_estimators}.
\input{Files/Tables/Parameters}

%% file: Files/Tables/Parameters.tex
\begin{table}[htbp]
\caption{Parameter values, JCBB algorithm block}
\centering
\begin{tabular}{p{1.5cm} p{3cm} p{2.5cm}}
\toprule
\textbf{Parameter} & \textbf{Explanation} & \textbf{Value} \\
\midrule
$\alpha$ & confidence level for $\chi$ test & 0.975 \\
$\gamma$ & ratio test threshold & 100.00 \\
$\mathbf{\Sigma_{e}}$ & state covariance & diag\{5,5,5,0.25,\\
& & 0.25,0.25\}*1e-2 \\
$\mathbf{\Sigma_{v}}$ & measurement covariance & diag\{50,50\}  \\
\bottomrule
\end{tabular}
\label{tab:parameters_JCBB}
\end{table}

\begin{table}[htbp]
\caption{Parameter values, estimators}
\centering
\begin{tabular}{p{1.5cm} p{3cm} p{2.5cm}}
\toprule
\textbf{Parameter} & \textbf{Explanation} & \textbf{Value} \\
\midrule
$\alpha_f$ & AEKF forget factor & 0.6 \\
$N_{p}$ & particle number & 1000 \\
$N_{eff}$ & effective particle number & 100 \\
$\mathbf{\Sigma_{e}}$ & state covariance & diag\{5,5,5,0.25,\\
& & 0.25,0.25\}*1e-6 \\
$\mathbf{\Sigma_{v}}$ & measurement covariance & diag\{25,25\}  \\
\bottomrule
\end{tabular}
\label{tab:parameters_estimators}
\end{table}